\documentclass[11pt]{article}
\usepackage[margin=1in]{geometry}
\usepackage[T1]{fontenc}
\usepackage{float}
\usepackage{booktabs,array,tabularx,amssymb,pifont,url}
\usepackage[numbers,sort]{natbib}
\usepackage[colorlinks=true,allcolors=blue]{hyperref}
\newcommand{\cmark}{\ding{51}}
\newcommand{\xmark}{\ding{55}}
\title{Typed Decision Models: An Early Evidence Audit and Evaluation Checklist}
\author{Lijuan Tang \qquad Yuemeng Zheng\\[4pt]Northeastern University}
\date{September 2026}
\begin{document}
\maketitle
\begin{abstract}
Typed decision models (TDMs) return probability distributions over caller-defined options without generating text. TypeSafe released Jev, a commercial typed decision model, on 15 September 2026, and a small body of evaluation and replication work appeared within days. We review 28 papers posted between 19 and 24 September and relate their findings to earlier work on label-probability classification, constrained decoding, reranking, calibration, and model cascades. In this early literature, the typed readout itself has not shown an independent accuracy advantage over comparable label-probability readouts. Jev's clearest gains are in latency and cost, while accuracy gaps remain on harder tasks. In practical deployments, confidence is often used to decide when to defer to a stronger model or a human. We use recurring weaknesses in these studies to derive a 14-item evaluation checklist for future TDM work. Because the evidence covers only the first nine days after the release of one hosted model, the review should be read as an early evidence map rather than a settled assessment of the model class.
\end{abstract}
\section{Introduction}
\label{sec:intro}

Many uses of models inside software are not conversational. A program may need to decide which queue a ticket belongs in, whether a retrieved passage answers a question, or whether a command is safe to run. In such settings, the desired output is often one of a few known values together with a score that downstream code can threshold.

Language models usually express these decisions as text. Structured prompting and constrained decoding can make the output easier to parse, but both still rely on token generation. Typed decision models (TDMs) expose a different interface: the caller supplies a state and one or more questions with finite option sets, and the model returns a probability distribution over exactly those options without producing free-form text. TypeSafe AI released Jev, a commercial TDM, on 15 September 2026 and presented it as a new class of ``System One'' model for fast, inexpensive decisions~\citep{typesafe2026jev}, a name taken from Kahneman's distinction between fast and slow thinking~\citep{kahneman2011thinking}.

The release prompted a rapid burst of research. Between 19 and 24 September 2026, our searches found 29 arXiv papers that mentioned or evaluated typed decision models; one was excluded because Jev appeared only in a renamed model-collection URL. The resulting 28-paper corpus spans service orchestration, agent control, memory, judging, clinical report evaluation, social-science annotation, video quality, game playing, security, robustness to context and injection, and open replications.

This paper asks two questions: what is technically new about the TDM interface, and what does the first week of evidence support? We read the 28 papers in full and recorded their comparators, headline accuracy, latency and cost results, calibration analyses, use of label log-probability baselines, statistical treatment, and handling of uncertainty. We also trace the main components of the interface to earlier work.

We treat this corpus as an early map of a fast-moving area rather than a mature survey of an established field. We define the interface and its lineage, document the public evidence on Jev and open TDMs, organize applications by the role the decision plays in the surrounding system, synthesize six recurring empirical patterns, and derive an evaluation checklist from failures already visible in the corpus. Jev's internals remain closed, so claims about mechanism apply only to open systems; hosted Jev is compared here at the level of its input-output contract.

Two concurrent working drafts, released on GitHub while this paper was in preparation, also review Jev and related models. \citet{luo2026jevsurvey} maintains an evidence survey of the early studies organized around calibration, selective control and open implementations, and \citet{anon2026decisions} place Jev within a broader history of machine-native decision models, from classical classifiers through structured prediction and learning to defer. Our scope is deliberately narrower. We audit the first wave of empirical TDM studies, ask which reported gains can be attributed to the typed readout itself, and derive an evaluation checklist, with corpus-level compliance counts, from recurring comparison and reporting failures.

\section{Definition and Lineage}
\label{sec:lineage}

\subsection{What we mean by a typed decision model}

A typed decision model takes a state s, which may be free text or a serialized data structure, and a set of questions Q = {q1, ..., qK}. Each question carries an instruction and a finite option set declared by the caller at request time, optionally with a rubric that defines what each option means. For every question, the model returns a probability distribution over the declared options and may also expose a scalar confidence. It returns no free-form text.

Three properties distinguish the interface from neighboring designs. First, support is closed at request time: the output distribution covers exactly the options supplied in the call. Unlike a conventional classifier, however, the label set need not be fixed at training time. Second, the interface is designed to answer multiple isolated questions about the same state in one forward execution, with each question evaluated without seeing the answers to the others~\citep{cheng2026thisthat,typesafe2026docs}. Third, the probability distribution is itself the product: downstream code is expected to threshold, rank, route, or defer on it rather than merely consume the argmax.

TypeSafe exposes this contract through Choice, Score, and Noul primitives~\citep{typesafe2026docs}. The open replications in our corpus primarily implement Choice, sometimes with Boolean or ordinal variants. We use ``TDM'' for systems that satisfy the three properties above and reserve ``Jev'' for TypeSafe's hosted model.

\begin{table}[t]
\centering
\small
\begin{tabularx}{\linewidth}{@{}>{\raggedright}p{2.6cm} >{\raggedright}p{4.2cm} >{\raggedright\arraybackslash}X@{}}
\toprule
\textbf{TDM component} & \textbf{Ancestor} & \textbf{What TDMs change} \\
\midrule
Scoring a label set by LM probability & Cloze/verbalizer classification \citep{gao2021lmbff,schick2021pet}; likelihood scoring \citep{brown2020gpt3} & Options declared per request rather than fixed verbalizers \\
Option-label readout & Multiple-choice symbol binding \citep{robinson2023mcqa} & Named options may include rubrics rather than letters \\
Guaranteed valid output & Grammar-constrained decoding \citep{geng2023gcd,willard2023outlines} & Returns a renormalized distribution without token generation \\
Fine-tuned label-token head & Yes/no rerankers \citep{nogueira2020monot5,zhang2025qwen3emb}; generative verifiers \citep{zhang2025genverifier} & Arbitrary option sets; many questions per call \\
Runtime labels in small encoders & NLI zero-shot classifiers \citep{laurer2023universal,yin2019benchmarking}; GLiClass \citep{stepanov2025gliclass} & Presented as a general decision interface \\
Rubric-conditioned judging & Prometheus \citep{kim2024prometheus}; G-Eval \citep{liu2023geval} & Rubric attached to each option \\
Confidence-based escalation & Reject option \citep{chow1970reject}; LLM cascades \citep{aggarwal2024automix,chen2024frugalgpt} & Cheap first stage with a native probability \\
Calibration training & RL with proper scoring rewards \citep{baniharouni2026doubt,damani2026rlcr} & Calibration is part of the product objective (RLCD) \\
\bottomrule
\end{tabularx}
\caption{Components of a TDM and their closest ancestors.}
\label{tab:lineage}
\end{table}

\subsection{Ancestors}

The interface draws directly on several earlier lines of work. Table~\ref{tab:lineage} summarizes the closest predecessors; the most relevant are label-probability classification, constrained output, discriminative judging, and confidence-based deferral.

Label-probability classification is the closest readout ancestor. Pattern-exploiting training recast classification as predicting one of a small set of verbalizer tokens~\citep{schick2021pet,schick2021small}, LM-BFF searched for suitable labels and templates~\citep{gao2021lmbff}, and GPT-3 popularized likelihood-based scoring without task-specific training~\citep{brown2020gpt3}. That literature also documented biases that matter directly for TDMs: probability mass can depend on label wording, option order, recency, and surface form~\citep{holtzman2021surface,min2022noisy,pezeshkpour2024order,zhao2021calibrate,zheng2024mcq,zhou2024batch}. In instruction-tuned models, first-token option probabilities can also disagree with the answer the model would generate~\citep{wang2024myanswer}. The option-name polarity failure measured by \citet{sun2026optionname} is therefore best understood as a new instance of a known family of label-sensitivity problems rather than as a failure unique to typed interfaces.

Constrained decoding provides a second lineage. Finite-state and grammar-constrained decoders already guarantee outputs that satisfy a schema~\citep{dong2025xgrammar,geng2023gcd,scholak2021picard,willard2023outlines}. A choice among n declared options is the simplest possible constrained grammar. TDMs differ mainly in returning a renormalized distribution directly rather than generating an allowed token sequence. Prior work also shows that masking can alter a model's probability distribution~\citep{beurerkellner2024domino,park2024gad} and that answer-only restrictions can reduce performance on tasks that benefit from explicit intermediate computation~\citep{banerjee2025crane,tam2024speak}.

The discriminative lineage is equally direct. Pointwise rerankers such as monoT5 score ``true'' versus ``false''~\citep{nogueira2020monot5}, Qwen3-Reranker reads a yes/no softmax~\citep{zhang2025qwen3emb}, reward models attach discriminative heads to language models~\citep{ouyang2022instructgpt}, often trained with pairwise objectives related to Bradley--Terry models~\citep{bradley1952rank}, generative verifiers use answer-token probabilities~\citep{zhang2025genverifier}, and rubric-conditioned judges such as G-Eval and Prometheus score outputs against explicit criteria~\citep{kim2024prometheus,liu2023geval}. Small encoders that score labels supplied at inference time also predate TDMs~\citep{laurer2023universal,stepanov2025gliclass,zaratiana2024gliner}. These systems show that a non-generative decision interface can be useful without implying that the underlying mechanism is novel.

TDM deployment also closely resembles selective classification and model cascading. Reject options, learning to defer, and confidence-based language-model cascades all predate Jev~\citep{aggarwal2024automix,chen2024frugalgpt,chow1970reject,geifman2017selective,gupta2024cascades,jitkrittum2023deferral,kamath2020selective,madras2018defer,mozannar2020defer,varshney2022cascading}. Temperature scaling and isotonic regression provide standard post-hoc calibration tools~\citep{guo2017calibration,zadrozny2002transforming}, while recent work has trained language models directly for calibrated confidence using proper scoring rewards~\citep{baniharouni2026doubt,damani2026rlcr}. At the observable interface level, TDMs primarily package caller-defined options, a type-level output guarantee, shared processing across multiple questions, and a probability intended for direct use by downstream software.

\section{Models and Readouts}
\label{sec:models}

\subsection{Jev: what is public}

Jev entered early access on 15 September 2026~\citep{typesafe2026jev}. Every hosted-Jev paper in our corpus that names a version uses 1.13 (\texttt{jev-1.13.0}); two record the snapshot \texttt{jev-1.13-20260917}~\citep{ibrahim2026css,li2026edge}. TypeSafe describes a new architecture, a parallel sampler, and a training method called Reinforcement Learning for Calibrated Decisions (RLCD), and states that Jev is ``neither small nor an LLM''~\citep{typesafe2026jev}. Architecture, training data, and reward are not public~\citep{ibrahim2026css}.

The published price is \$0.042 per million input tokens with no output charge, and the vendor reports end-to-end response times of 70--500\,ms~\citep{typesafe2026jev}. Options per question are capped at 255, and for high-cardinality choices the vendor describes a two-stage procedure that first scores options independently and then makes an explicit choice~\citep{typesafe2026jev}. TypeSafe does not publish results on public benchmarks and instead reports workflow evaluations against the average answer of two frontier models~\citep{typesafe2026jev}.

Independent studies add several observable properties. Jev is not deterministic: 4 of 510 labels changed on an identical repeat~\citep{zhang2026contract}, and \citet{sun2026optionname} report that at most 1.33\% of decisions change between two identical runs on 300 questions. Its probabilities lie on a two-decimal grid from 0.01 to 1.00, with no zero observed in 2.4 million answers~\citep{rafe2026crash}. On one population-experiment workload, asking for one Noul probability per option and renormalizing was much better calibrated than asking for a relative Choice distribution~\citep{li2026kite}. Because the internals are closed, the evidence reviewed here cannot confirm or refute the vendor's architectural claims.

\subsection{Open TDMs}

The corpus names at least fifteen open systems that expose a similar interface, although only a subset documents the readout. Table~\ref{tab:open} lists the systems for which the mapping from hidden representations or logits to option probabilities is described. The label ``Open-Jev'' is used for different artifacts across papers; here it refers only to the third-party DeBERTa-v3-large checkpoint evaluated by \citet{sun2026optionname}, who also use the name for a separate dataset.

\begin{table}[t]
\centering
\small
\begin{tabularx}{\linewidth}{@{}>{\raggedright}p{2.5cm} >{\raggedright}p{2.8cm} >{\raggedright}X >{\raggedright\arraybackslash}p{2.6cm}@{}}
\toprule
\textbf{Model} & \textbf{Backbone} & \textbf{Readout} & \textbf{Training} \\
\midrule
this-that-model-1.0 \citep{cheng2026thisthat} & 1.88B hybrid linear-attention decoder & Answer-slot hidden state scored against output-embedding rows of single-token option labels; restricted softmax & Full fine-tune; cross-entropy + Brier \\
JevLite \citep{ren2026callscreen} & Qwen3-4B, LoRA & Softmax over allowed label-token logits & Cross-entropy + Brier; temperature scaling \\
Visual Jev \citep{yu2026visualjev} & Qwen3-VL-4B/8B, LoRA & LM-head logits of option letters, renormalized over candidates & Answer SFT; dedicated typed head tried and dropped \\
SemIf-4B \citep{ibrahim2026css} & Qwen3.5-4B, frozen & Option-letter logits & None \\
Qwen3-0.6B-RLCD \citep{ibrahim2026css} & Qwen3-0.6B-Base & Letter-token readout & RL for calibrated decisions (third-party replication) \\
PixelJev \citep{zhou2026pixeljev} & Qwen3.5-2B/4B, optional LoRA & Option-letter logits at the first answer position, restricted to the candidate slots & Few-shot LoRA; temperature fitted on held-out data \\
Laya \citep{sun2026optionname} & ModernBERT-large & Contextual embedding of a mask marker placed at each option & n.r. \\
Open-Jev \citep{sun2026optionname} & DeBERTa-v3-large & Mean representation over option name and definition & n.r. \\
\bottomrule
\end{tabularx}
\caption{Open TDMs with documented readouts. n.r.: not reported in the corpus.}
\label{tab:open}
\end{table}

\subsection{Readout geometries and serving}

The documented readouts fall into three broad geometries. Decoder models restrict the language-model softmax to allowed answer tokens at a chosen answer position. Laya scores a marker token associated with each option. Open-Jev pools representations over the option name and definition before scoring. So far, these differences are more evident in robustness than in headline accuracy.

Two comparisons are especially informative. \citet{yu2026visualjev} find that a dedicated typed head does not improve macro accuracy over the ordinary LM-head readout when backbone, data, and training budget are matched; a fixed-slot head trained on at most four options also degrades sharply on eight. For yes/no option names, \citet{sun2026optionname} find that a DeBERTa model pooling over the full option text flips about four times less often than a ModernBERT model scoring a single marker token (their Table 6: 19.5\% vs.\ 76.9\%); the authors suggest, without isolating it, that pooling may explain part of the gap. Readout design may therefore affect sensitivity to labels even when average accuracy is unchanged.

Observed speed advantages can partly reflect serving design rather than the readout alone. Shared encoding of a common state can amortize work across many questions. In Visual Jev, sharing the image prefix and batching the question suffixes cuts amortized time per question 8.9-fold relative to serial execution and 3.4-fold relative to batching without prefix reuse at 32 questions per image, at the cost of higher peak memory; a single question is slower~\citep{yu2026visualjev}. For hosted Jev, requesting all 17 labels of a contract at once rather than one raises median latency only from 1.15 to 1.22\,s, although the authors caution that network and serving overhead may explain this~\citep{zhang2026contract}. In principle, the same systems-level optimization could be applied to other label-probability readouts.

\subsection{Training objectives}

The open models use several training objectives rather than a single recipe. Two combine cross-entropy with a Brier term~\citep{cheng2026thisthat,ren2026callscreen}, one uses ordinary supervised fine-tuning on answers~\citep{yu2026visualjev}, and \citet{ibrahim2026css} evaluate a third-party 0.6B replication of an RLCD-style objective. That replication improves macro-F1 by about eight points over its untrained base, but the gain is smaller than the gap to larger untrained backbones in the same comparison.

The most controlled comparison comes from \citet{ren2026callscreen}: a Qwen3-4B model fine-tuned to generate its answer and then read through label-token probabilities reaches AUROC .950, against .972 for a matched single-seed typed-readout model, with the same calibration error (ECE .050) after temperature scaling; against the three-seed typed ensemble (.974) the difference is not significant. Taken together with \citet{yu2026visualjev}, the current evidence suggests that task training and backbone quality matter more for accuracy than the choice between a typed head and a conventional label-probability readout.

\subsection{Open questions about the models}

Several model-level questions remain unresolved. Public evidence cannot separate the effects of Jev's scale, data, RLCD objective, and architecture. Every hosted study covers essentially the same model version within one week, so cross-version stability is unknown. No corpus paper tests the two-stage procedure the vendor describes for high-cardinality choices, leaving its effect on ranking and calibration unmeasured.

\section{Where the Decision Sits}
\label{sec:taxonomy}

The same decision interface plays different roles in different systems. We organize applications by what the TDM controls and by what happens when confidence is low (Table~\ref{tab:taxonomy}). This distinction matters because an error in a one-shot pipeline step, a repeated controller decision, a judge verdict, and a measurement instrument can have very different consequences.

\begin{table}[t]
\centering
\small
\begin{tabularx}{\linewidth}{@{}>{\raggedright}p{2.6cm} >{\raggedright}X >{\raggedright\arraybackslash}p{3.6cm}@{}}
\toprule
\textbf{Role} & \textbf{Papers} & \textbf{Where low confidence goes} \\
\midrule
Step replacement & Edge/6G orchestration \citep{li2026edge,li2026sixg}; scientific choices \citep{deng2026science}; contract inference \citep{zhang2026contract}; video quality \citep{robitza2026jevqa}; standalone visual and grid-world questions \citep{yu2026visualjev,zhou2026pixeljev,cheng2026thisthat}; numerical decoding \citep{ye2026numericjev}; mobile GUI actions \citep{zhang2026jevmobile}; edge triage \citep{dagli2026fractal} & Planner when the TDM selects a done/blocked option \citep{zhang2026jevmobile}; not reported in the other studies \\
Controller & Tool-using agent \citep{wu2026reflex}; memory control \citep{jiang2026jevmem}; game control \citep{ma2026jevstar}; coding-agent routing \citep{abbasi2026harness} & Strong LLM by confidence \citep{abbasi2026harness,wu2026reflex}; planner by schedule \citep{ma2026jevstar}; not reported in the other studies \\
Judge or screener & Preference/factuality judging \citep{li2026judge}; rubric judging \citep{rao2026rubricjudge}; radiology reports \citep{huang2026radiology}; scam calls \citep{ren2026callscreen}; alignment-failure detection \citep{guo2026justaskjev}; security findings \citep{barbosa2026pentest} & Stronger judge by confidence \citep{li2026judge}; LLM judge by confidence, replayed \citep{rao2026rubricjudge}; human review of needs-review findings by design, not exercised in the evaluation \citep{barbosa2026pentest}; not reported in the other studies \\
Measurement instrument & Social-science annotation \citep{ibrahim2026css}; crash narratives \citep{rafe2026crash}; population simulation \citep{li2026kite} & LLM \citep{ibrahim2026css} or human \citep{rafe2026crash} by confidence; flagship model on fixed sample \citep{li2026kite} \\
\midrule
Diagnostic & Option-name polarity \citep{sun2026optionname}; context flipping \citep{xu2026jevout}; prompt injection \citep{wu2026hijacking} & -- \\
Ecosystem & Public GitHub projects \citep{ling2026wild} & -- \\
\bottomrule
\end{tabularx}
\caption{Corpus papers by the role of the TDM and where low-confidence decisions go.}
\label{tab:taxonomy}
\end{table}

\paragraph{Step replacement.}
In these systems, the TDM replaces one model call inside an otherwise fixed pipeline: an intent field, a semantic relation, or a quality score. The central question is whether it preserves task performance while reducing latency or cost. Several parity claims come from this group, but the corresponding tests are often near ceiling~\citep{deng2026science,li2026sixg}, as is the first of two studies in \citet{li2026edge}, and some reported gains shrink or disappear once decisions are cached~\citep{li2026edge,li2026sixg}.

\paragraph{Controller.}
Here a sequence of TDM decisions changes state: whether to retrieve again, which tool to call, or which game action to take. Errors can compound, so per-step accuracy is a weak proxy for end-to-end success. In the corpus, deciding whether to act is substantially harder than choosing among actions~\citep{wu2026reflex}, and in one full game a per-step chooser without a planner never saved resources for a later expansion, although this was not a controlled ablation~\citep{ma2026jevstar}.

\paragraph{Judge or screener.}
A TDM can cheaply screen outputs produced elsewhere, making escalation attractive: routine preference and grounded-factuality cases can be handled cheaply, while difficult cases are passed to a stronger judge. The difficulty is that confidence becomes less informative precisely on adversarial and reference-free examples~\citep{li2026judge}, and that the fallback judge may repeat the TDM's confident errors, leaving little for escalation to recover~\citep{rao2026rubricjudge}.

\paragraph{Measurement instrument.}
Here TDM outputs become data for downstream analysis, as in crash coding, social-science annotation, or population simulation. Aggregate accuracy alone is insufficient; systematic error, prevalence-specific calibration, run-to-run stability, and downstream robustness also matter~\citep{baumann2025hacking,egami2023dsl}. The strongest measurement papers in the corpus explicitly audit calibration, preregister analyses, or derive human-review budgets~\citep{ibrahim2026css,li2026kite,rafe2026crash}.

Across these roles, the treatment of uncertainty distinguishes ordinary classification from deployments that make explicit use of confidence. If low-confidence cases go nowhere, the TDM should be compared with a conventional classifier. If they go to a stronger model, the relevant baseline is a cheap generative cascade. If they go to humans, the evaluation should report held-out calibration and an explicit review budget.

\section{What the Evidence Shows}
\label{sec:evidence}

\paragraph{Corpus construction.}
We searched arXiv on 24 September 2026 and repeated the search on 25 September, using the API query \texttt{all:jev} and abstract searches for ``typed decision model'', ``decision model'', ``System One model'' and ``TypeSafe''. By 25 September the \texttt{all:jev} query returned 28 records, one of them an unrelated 2024 economics paper by an author named Jev; the abstract searches added two papers whose searchable metadata mention Jev only inside the compound ``JevBench'' or not at all~\citep{dagli2026fractal,ibrahim2026css}. We included papers that evaluate hosted Jev or build or evaluate a model with the typed-decision interface defined in Section~\ref{sec:lineage}, and excluded papers that mention Jev only incidentally. This removed one reranker report whose only reference to Jev is a renamed model-collection URL, leaving 28 papers. Blog posts, leaderboards and model cards were used only as sources of vendor facts, not as corpus items. All quantitative claims were extracted from the latest arXiv version available at extraction (24 September for the first 17 papers, 25 September for the remaining 11); versions are listed in Appendix~\ref{app:versions}. New papers continued to appear at several per day during the review, so the corpus is a snapshot.

\paragraph{Extraction and coding.}
Each paper was read in full and coded against a fixed extraction schema covering comparators, headline accuracy, latency and cost with their source locations, calibration, label-probability baselines, escalation, statistics, and validity threats. Extraction was assisted by LLM-based tools. A second pass, run without access to the first, re-checked every quantitative claim used in the tables and findings against the source full text; it led to wording corrections and two factual corrections, one of attribution and one of a denominator, all of which were applied. Checklist items were coded in two steps, applicability and then compliance, using the criteria in Appendix~\ref{app:coding}; cases that could not be determined from the text are marked unclear, and five items that concern disclosures we could not score reliably are left unscored. The authors reviewed the resulting coding and the verification results.

Our corpus contains 28 papers posted between 19 and 24 September 2026: 22 evaluate hosted Jev, five build open or alternative TDMs, one of them without reference to TypeSafe~\citep{yu2026visualjev}, and one analyzes public GitHub projects that use Jev~\citep{ling2026wild}. Tables~\ref{tab:evidence} and~\ref{tab:evidence2} summarize the headline comparisons; the ecosystem study has none and is omitted. Six patterns recur across the corpus.

\begin{table*}[!tp]
\centering
\scriptsize
\setlength{\tabcolsep}{3pt}
\begin{tabularx}{\textwidth}{@{}>{\raggedright}p{2.3cm} >{\raggedright}p{2.4cm} >{\raggedright\arraybackslash}X >{\raggedright}p{1.7cm} >{\raggedright}p{1.9cm} >{\centering}p{1.0cm} >{\centering}p{0.8cm} >{\raggedright\arraybackslash}p{1.5cm}@{}}
\toprule
\textbf{Paper} & \textbf{Comparator / control} & \textbf{Accuracy (TDM vs.\ comparator)} & \textbf{Latency} & \textbf{Cost} & \textbf{Calib.} & \textbf{LP} & \textbf{Stats} \\
\midrule
\multicolumn{8}{@{}l}{\emph{Open and alternative TDMs}} \\
\citet{cheng2026thisthat} & GPT-5.6 & 0.844 vs.\ 0.897 (own bench); 0.941 vs.\ Jev 0.765 (68 items) & 30.9\,ms local & local (electricity) & Yes & ? & none \\
\citet{ren2026callscreen} & same backbone, label token & AUROC .974 vs.\ .950 (n.s.) & 64.5\,ms & n.r. & Yes & Yes & bootstrap, 3 seeds \\
\citet{yu2026visualjev} & typed-head control (default: LM head) & macro acc.\ .761 vs.\ .761 & 8.9$\times$ faster$^{*}$ & n.r. & appendix & Yes & 3 seeds (SD); bootstrap described, CIs not reported \\
\citet{zhou2026pixeljev} & same backbone, greedy generation & readout = generation when valid; +1.88\,pp (adapted, ScienceQA) only from invalid generations & 1.3--1.4$\times$ faster than gen.\ (1.03$\times$ vs.\ one-token constrained gen.) & n.r. & Yes & Yes & group bootstrap, 3 seeds \\
\citet{dagli2026fractal} & internal ablations; random & JevBench 55.4\% vs.\ 31.8\% random (v2, self-reported) & $\approx$2--9\,ms (inconsistent) & estimates only & No & No & CIs on in-house tests only \\
\midrule
\multicolumn{8}{@{}l}{\emph{Step replacement}} \\
\citet{li2026edge} & DeepSeek V4.1 Flash & 212--214 vs.\ 215--216 of 216 & 16--27\% lower$^{**}$ & 69--73\% lower & No & No & none \\
\citet{li2026sixg} & DeepSeek, Gemini & 63/63 vs.\ 63/63 (ceiling) & 22\% / 62\% lower & 1.7$\times$ more (DeepSeek); 4.6$\times$ less (Gemini) & No & No & none \\
\citet{deng2026science} & eleven LLM configurations (five tie) & 100/100 vs.\ 100/100 (20 items) & 0.34 vs.\ 1.39\,s (GPT-5.6 Sol) & $\approx$50$\times$ less (vs.\ GPT-5.6 Sol) & No & No & 5 repeats \\
\citet{robitza2026jevqa} & ITU-T P.1204.1 & PLCC .737 vs.\ .733 (metadata); bitstream .797 vs.\ trained $\approx$.96 & n.r.\ per call & \$4 total & partial & No & none \\
\citet{zhang2026contract} & Gemini 3.1 Pro & 77.4\% vs.\ 83.2\% & 1.24\,s & $\approx$37$\times$ less & No & No & contract bootstrap \\
\citet{ye2026numericjev} & direct 100-way choice (same Jev) & MAPE 19.7 vs.\ 21.7 (n.s.); range-normalized error 1.84 vs.\ 5.18 & n.r. & \$0.29 fixed-grid evaluation & No & No & family bootstrap \\
\citet{zhang2026jevmobile} & step-wise VLM & success .79 vs.\ .84 & 133 vs.\ 197\,s per success & 73\% lower per success & No & No & none \\
\midrule
\multicolumn{8}{@{}l}{\emph{Controllers}} \\
\citet{jiang2026jevmem} & MAGMA & judge score .777 vs.\ .700 & 0.93 vs.\ 1.47\,s & n.r. & No & No & none \\
\citet{ma2026jevstar} & Jev-only controller (not matched) & 4/4 wins (Lv5--7) vs.\ 0/1 (Lv2, time limit) & 0.42\,s & \$0.15 of \$3.71/game & No & No & none \\
\citet{wu2026reflex} & cheap generative cascade & $\tau^2$: .850 vs.\ .917 (n.s.) & n.r. & \$0.057 vs.\ \$0.041 & Yes & No & bootstrap, 1 run \\
\citet{abbasi2026harness} & no routing; oracle labels & labels 63--82\% vs.\ author tags ($n\approx109$) & not measured & 13.8\% lower (simulated) & No & No & none \\
\bottomrule
\end{tabularx}
\caption{The TDM corpus, part 1: open and alternative TDMs, step replacement and controllers (arXiv, 19--24 Sept.\ 2026). \emph{Calib.}: TDM probabilities evaluated for calibration. \emph{LP}: at least one baseline reads label log-probabilities from an ordinary or same-backbone LM; ?: unclear. n.r.: not reported. $^{*}$Amortized time per question at 32 questions per image versus serial execution. $^{**}$Median decision latency; the median paired gap falls below 1\,ms once repeated requests are cached.}
\label{tab:evidence}
\end{table*}

\begin{table*}[!tp]
\centering
\scriptsize
\setlength{\tabcolsep}{3pt}
\begin{tabularx}{\textwidth}{@{}>{\raggedright}p{2.3cm} >{\raggedright}p{2.4cm} >{\raggedright\arraybackslash}X >{\raggedright}p{1.7cm} >{\raggedright}p{1.9cm} >{\centering}p{1.0cm} >{\centering}p{0.8cm} >{\raggedright\arraybackslash}p{1.5cm}@{}}
\toprule
\textbf{Paper} & \textbf{Comparator / control} & \textbf{Accuracy (TDM vs.\ comparator)} & \textbf{Latency} & \textbf{Cost} & \textbf{Calib.} & \textbf{LP} & \textbf{Stats} \\
\midrule
\multicolumn{8}{@{}l}{\emph{Judges and screeners}} \\
\citet{li2026judge} & GPT-6 Astra & RewardBench $-$1.3; JudgeBench $-$14.6; RM-Bench hard $-$19.8 & 0.15 vs.\ 1.89\,s & 0.36\% of fee & Yes & No & cluster bootstrap \\
\citet{rao2026rubricjudge} & three flash-tier LLM judges & differs in 8/27 comparisons (4 ahead, 4 behind; graded up to $-$16.4) & 0.18--0.20\,s & 29--325$\times$ less & Yes & No & bootstrap, Holm \\
\citet{huang2026radiology} & RadMatch (local) & $\tau_b$ .573 vs.\ .580; sig.\ errors $-$.145; 200-pair RadEvalExpert subset .322 vs.\ .453 & 434\,ms/req.\ (median, Full) & 2.3--2.6 cents/100 pairs & Yes & No & bootstrap; 5-run repeatability on 50 pairs \\
\citet{guo2026justaskjev} & in-domain TF-IDF classifier (accuracy); LLM judges (cost) & median AUROC .886 vs.\ .754 & 0.31\,s & 63$\times$ less at list prices; 6$\times$ under repricing & Yes & No & bootstrap \\
\citet{barbosa2026pentest} & same harness without TDM & 9/13 vs.\ 10/13 scenarios (1 run each) & 26.9 vs.\ 32.2\,min/run & $<$\$5 TypeSafe API & No & No & none ($n=1$ per arm) \\
\midrule
\multicolumn{8}{@{}l}{\emph{Measurement instruments}} \\
\citet{ibrahim2026css} & per-task best of 19 LLMs & macro-F1 $-$11.6 median; behind on 14/15 & n.r. & 44$\times$ less & Yes & open only & prereg., BH-FDR \\
\citet{rafe2026crash} & Claude Fable 5.1 & F1 .908 vs.\ .967 (sig.) & 0.20\,s & 26--129$\times$ less$^{***}$ & Yes & No & cluster bootstrap \\
\citet{li2026kite} & GPT-6 Astra & decision gain .272 vs.\ .336 (n.s.) & 213\,ms & 312--455$\times$ less & Yes & No & bootstrap, prereg. \\
\midrule
\multicolumn{8}{@{}l}{\emph{Diagnostics}} \\
\citet{sun2026optionname} & neutral-name reassignment; Laya, Open-Jev & AUC .815 $\to$ .581 under polarity swap & -- & -- & -- & -- & bootstrap \\
\citet{xu2026jevout} & open TDMs; same-backbone label scorer & context flips 61.4\% of correct decisions & n.r. & n.r. & No & Yes & Wilson CIs, bootstrap \\
\citet{wu2026hijacking} & attack variants (no model baseline) & attack success 1.8\% $\to$ 3.5\% (adaptive) & n.r. & n.r. & No & No & bootstrap, registered hypotheses (DH-2/3) \\
\bottomrule
\end{tabularx}
\caption{The TDM corpus, part 2: judges and screeners, measurement instruments and diagnostics. \emph{Calib.}: TDM probabilities evaluated for calibration. \emph{LP}: at least one baseline reads label log-probabilities from an ordinary or same-backbone LM. n.r.: not reported. $^{***}$Against list-price scenarios for frontier models, which were not metered.}
\label{tab:evidence2}
\end{table*}

\paragraph{Finding 1: the typed readout has not yet shown an independent accuracy advantage.}
Twenty-one of the 27 papers that evaluate a model include no label-probability baseline, and one more is unclear because it scores hosted models' token probabilities without stating how they were extracted~\citep{cheng2026thisthat}; where they compare against other models at all, those models generate their answer, usually as a JSON label or a verbalized probability. This mirrors the vendor's own comparison adapter, which asks LLMs to state probabilities rather than exposing token log-probabilities~\citep{typesafe2026adapter}. The five papers that do examine label probabilities from an ordinary or same-backbone model point in the same direction. \citet{ren2026callscreen} report AUROC .950 for the label-token readout of a generative fine-tune against .974 for a three-seed typed ensemble, a difference that is not significant; a matched single-seed typed model reaches .972. \citet{yu2026visualjev} obtain the same rounded .761 macro accuracy (overlapping seed ranges) from a dedicated typed head and the ordinary LM head. \citet{zhou2026pixeljev} find that, with weights fixed, restricted option-logit readout matches greedy generation exactly whenever the generated answer is valid, so its only gain comes from invalid generations. \citet{xu2026jevout} compare a Jev-style open decision interface with a plain label-probability scorer on the same frozen weights: the typed interface is not more robust to answer-preserving context (flip rates 72.6\% vs.\ 64.9\%, measured on different item sets), and its clean-accuracy advantage is concentrated in tool routing. \citet{ibrahim2026css} find that an untrained 4B model read through option-letter logits is the open decision system with the highest median macro-F1 in their comparison. These results do not establish that all readouts are equivalent, but they show why accuracy gains cannot be attributed to the typed interface without a matched label-probability baseline.

\paragraph{Finding 2: reported parity often occurs on saturated tests.}
Hosted Jev trails stronger hosted models on several discriminating evaluations: by a median 11.6 macro-F1 points in social-science annotation on the benchmark of \citet{ziems2024css}~\citep{ibrahim2026css}, 5.8 accuracy points on ContractNLI~\citep{zhang2026contract}, 14.6 points on JudgeBench~\citep{tan2025judgebench} and 19.8 on hard RM-Bench~\citep{liu2025rmbench} pairs~\citep{li2026judge}, 0.059 F1 on crash-narrative coding~\citep{rafe2026crash}, and up to 16.4 points on graded rubric criteria, although it is never significantly behind flash-tier LLM judges on binary criteria~\citep{rao2026rubricjudge}. By contrast, reported parity often occurs where nearly every strong model is at ceiling: 100/100 answers on 20 scientific choices~\citep{deng2026science}, 63/63 contract checks~\citep{li2026sixg}, or 212--216 of 216 requests with all intent fields correct~\citep{li2026edge}. These ceiling-level results do not establish model equivalence. The same corpus also shows weaknesses on multi-step arithmetic~\citep{cheng2026thisthat}, deciding whether to call a tool~\citep{wu2026reflex}, long-horizon control (in a single game, confounded by an affordability filter)~\citep{ma2026jevstar}, and simulated interventions~\citep{li2026kite}. One possible explanation is that the TDM interface provides no explicit iterative or generated intermediate computation, a setting in which earlier work has also found answer-only restrictions costly~\citep{tam2024speak}. Decisions are also sensitive to context: optimized answer-preserving additions redirect 61.4\% of Jev's initially correct decisions, against 16.9\% for a single target-aware addition and 2.2\% for a neutral one~\citep{xu2026jevout}, and score-guided prompt injection raises attack success from 1.8\% to 3.5\%~\citep{wu2026hijacking}.

\paragraph{Finding 3: latency and cost gains are substantial but context-dependent.}
Most reported median response times for single hosted-Jev requests fall between about 0.15 and 0.45 seconds~\citep{li2026judge,ma2026jevstar,rafe2026crash}, and several studies report fees tens to hundreds of times below frontier models~\citep{ibrahim2026css,li2026kite,li2026judge}. Later studies repeat this pattern with qualifications: rubric judging runs 30--220 times faster and 29--325 times cheaper than flash-tier LLM judges~\citep{rao2026rubricjudge}; a mobile agent's 73\% cost reduction is attributed to its delegated workflow (fewer VLM decisions), while the TDM's own charges are negligible and its contribution is not isolated~\citep{zhang2026jevmobile}; and one routing study's 13.8\% saving is simulated at list prices, with latency taken from vendor figures~\citep{abbasi2026harness}. In edge orchestration, however, repeated-request caching shrinks the median paired latency difference to under a millisecond~\citep{li2026edge}. In the 6G study, most of the completion gain comes from caching interpretation and delegating placement to a numerical scheduler (Jev rises from 43\% to 96\% completion), which dwarfs the 3.5--8.4-point gap between models~\citep{li2026sixg}. The same research group reports Jev as 70\% cheaper than DeepSeek in one paper and 1.7 times more expensive per call in another, on different workloads~\citep{li2026edge,li2026sixg}. Seven papers that evaluate a model report no latency at all.

\paragraph{Finding 4: calibration must be evaluated per model and workload.}
Jev's probabilities often rank examples usefully, but raw calibration is inconsistent. Against 2,416 human judgments in a crash-coding study, probabilities overstate prevalence (4.8\% vs.\ 2.7\%) and have a calibration slope of 1.63; a two-decimal output grid also limits resolution for labels rarer than 1\%~\citep{rafe2026crash}. Held-out recalibration reduces calibration error by a factor of 3.3 in that study. \citet{ibrahim2026css} find Jev better calibrated than most LLMs' verbalized confidence but worse than the best three, while a single temperature parameter is enough for some cheap LLMs to overtake it. Temperatures fitted on small pilot sets transfer unevenly even to held-out items of the same judging workload, and the fitted values differ widely across workloads~\citep{li2026judge}. High confidence also does not eliminate serious errors: on one empathy task, Jev is highly confident while performing near base rate~\citep{ibrahim2026css}, and wrong tool calls on BFCL have mean confidence .778~\citep{wu2026reflex}. On alignment-failure detection, pooled ECE is 0.047 but median per-benchmark ECE is 0.168, because the mean probability misses each benchmark's base rate~\citep{guo2026justaskjev}; in rubric judging, confidence ranks errors only weakly on graded panels (AUROC 0.57--0.70) and at chance on one~\citep{rao2026rubricjudge}; and in an exploratory numerical-decoding experiment, Jev's threshold probabilities are non-monotone in 25 of 48 query groups~\citep{ye2026numericjev}. These patterns are consistent with earlier findings on token-level and verbalized confidence in language models~\citep{kadavath2022know,tian2023justask,xiong2024express}.

\paragraph{Finding 5: confidence-based escalation is a recurring deployment pattern.}
Across judging, annotation, agent control, and data coding, the recurring design is to accept a TDM answer above a threshold and send the remainder to a stronger model or a human. The reported savings are often meaningful: a frozen judge cascade reaches 92.5 versus GPT-6's 93.1 at 57\% of the fee~\citep{li2026judge}; annotation cascades match the LLM alone at 27--56\% of its cost~\citep{ibrahim2026css}; an agent cascade makes 72.7\% fewer strong-model calls~\citep{wu2026reflex}; and a pooled review budget of 0.8\% of flagged crash records reaches 90\% precision against human labels, although per-variable budgets range up to 72\%~\citep{rafe2026crash}. The pattern also has clear limits. Thresholds chosen on a selection set do not hold up on held-out items for every fallback model~\citep{li2026judge}, confidence is much less informative on adversarial or reference-free inputs~\citep{li2026judge}, when the stronger model shares the TDM's errors escalation recovers little (in rubric judging, LLM judges repeat Jev's wrong answer in 96\% of their verdicts on its 12 most confident errors per graded panel (84 pairs), and a replayed cross-fitted cascade gains at most 1.5 points on average over the best single judge, 2.0 with oracle thresholds~\citep{rao2026rubricjudge}), and in the one study that compares against a cheap generative cascade, the TDM cascade wins on the authors' own benchmark but is more expensive, and not more accurate, on the external $\tau^2$-bench, where the authors retire their superiority claim~\citep{wu2026reflex}. The benefit of escalation therefore depends on how complementary the errors of the two stages are.

\paragraph{Finding 6: reporting and evaluation practices vary substantially across the corpus.}
Twelve of the 27 papers that evaluate a model fail our R2 criterion of reporting confidence intervals or significance tests on headline results. At least seven make post-evaluation design, recipe, metric, or threshold choices~\citep{dagli2026fractal,li2026kite,rao2026rubricjudge,ren2026callscreen,robitza2026jevqa,xu2026jevout,ye2026numericjev}, including a sequence of input redesigns that moves video-quality PLCC from .576 to .797~\citep{robitza2026jevqa} and a headline metric added after the original evaluation~\citep{ye2026numericjev}; separately, three add comparators after seeing results~\citep{dagli2026fractal,li2026judge,zhang2026contract}. At least seven use synthetic data in at least part of their evaluation~\citep{abbasi2026harness,cheng2026thisthat,li2026edge,ren2026callscreen,wu2026hijacking,wu2026reflex,ye2026numericjev}, including one setting in which an agent generates both text and gold labels~\citep{li2026edge}. The hosted model is also nondeterministic~\citep{rao2026rubricjudge,sun2026optionname,wu2026hijacking,zhang2026contract}. In the versions we reviewed, at least three papers contain internal numerical inconsistencies~\citep{cheng2026thisthat,dagli2026fractal,jiang2026jevmem}, and two sibling papers report opposite cost directions for the same comparator on different workloads~\citep{li2026edge,li2026sixg}. The thinnest evidence comes from a single run per condition~\citep{barbosa2026pentest} and from results self-reported on the authors' own infrastructure, where, in the paper's second version, 100\% output-format compliance sits alongside 55.4\% accuracy~\citep{dagli2026fractal}. Given the compressed publication window, these issues should not be read as unique to TDM research; instead, they motivate concrete reporting requirements and caution against broad claims about the model class.

\section{An Evaluation Checklist for Typed Decision Models}
\label{sec:checklist}

Table~\ref{tab:checklist} turns the recurring failures above into a compact checklist. For each scored item we report compliance among the papers to which the item applies rather than among all 28, so that non-compliance, non-applicability and insufficient reporting stay distinct; Appendix~\ref{app:coding} gives the criteria and the per-paper coding. The aim is not to propose a new evaluation theory but to make comparisons easier to interpret and reproduce. Several of the most important items are also inexpensive: include a label log-probability baseline when weights or logprobs are available; report the spread of performance across models so ceiling effects are visible; repeat hosted-model runs; evaluate calibration on held-out data; and freeze cascade thresholds before testing transfer.

\begin{table}[H]
\centering
\scriptsize
\renewcommand{\arraystretch}{0.9}
\setlength{\tabcolsep}{3pt}
\begin{tabularx}{\textwidth}{@{}r >{\raggedright}X >{\raggedright\arraybackslash}X >{\centering\arraybackslash}p{1.45cm}@{}}
\toprule
\textbf{ID} & \textbf{Report} & \textbf{Why it matters / failure observed} & \textbf{Count} \\
\midrule
\multicolumn{4}{@{}l}{\emph{Baselines}} \\
B1 & A baseline that reads label log-probabilities from an ordinary or same-backbone LM. & Otherwise interface gains cannot be separated from model gains; the same backbone is the most informative choice; where tested, the readout added little accuracy (Finding 1). & 5/26; 1 unclear \\
B2 & Where a confidence threshold passes decisions to another model: a cheap generative cascade as a baseline. & On external $\tau^2$-bench a cheap generative cascade was cheaper than the TDM cascade and not less accurate~\citep{wu2026reflex}; two of six open-weight LLMs in Jev's price range had higher median macro-F1~\citep{ibrahim2026css}. & 1/5 \\
B3 & Where labeled data suffice to train one: a task-specific model (encoder, classical classifier, reward model or probe) as a baseline. & Fine-tuned encoders were not significantly worse than an open TDM in ranking and had nominally lower ECE (but higher Brier) and were faster~\citep{ren2026callscreen}; trained quality models stayed well ahead~\citep{robitza2026jevqa}. & 7/13; 3 unclear \\
\midrule
\multicolumn{4}{@{}l}{\emph{Test design}} \\
T1 & Spread between models on the test; avoid parity claims on saturated tests. & Several parity results come from tests on which strong models are at or near ceiling (Finding 2). & -- \\
T2 & Correctness or agreement of individual decisions across repeated, reordered or reworded requests, not only aggregate accuracy. & Equal aggregate scores hid different decisions~\citep{zhang2026contract}; correct final labels hid wrong intermediate choices~\citep{deng2026science}. & 17/26; 1 unclear \\
T3 & Option-name invariance test: swap names across rubrics or compare neutral with semantic names. & Swapping no/yes across rubrics moved Jev AUC from .815 to .581 with zero type errors~\citep{sun2026optionname}; asking through Score instead of Choice moved graded accuracy by up to 10.5 points~\citep{rao2026rubricjudge}. & 1/27 \\
T4 & Data provenance: synthetic or real, gold-label source, and whether a public benchmark predates the model. & The corpus includes agent-generated texts with agent-generated labels~\citep{li2026edge} and unresolved contamination concerns~\citep{ibrahim2026css,zhang2026contract}. & -- \\
\midrule
\multicolumn{4}{@{}l}{\emph{Reproducibility and statistics}} \\
R1 & Exact model version and run date. & Hosted TDMs are closed and may change under a stable name; several papers provide neither. & -- \\
R2 & Confidence intervals or significance tests on headline results. & Twelve of the 27 report neither for their headline results. Because the hosted model is not deterministic~\citep{sun2026optionname,zhang2026contract}, repeated-run stability should also be reported where feasible. & 15/27 \\
R3 & Disclose prompts, recipes, baselines, or thresholds chosen after seeing test data. & At least seven papers make such choices; one series of redesigns moves the headline metric from .576 to .797~\citep{robitza2026jevqa}. & -- \\
\midrule
\multicolumn{4}{@{}l}{\emph{Confidence and deployment}} \\
C1 & Where the TDM's probabilities are used or described as calibrated: calibration measured with ECE, Brier score, reliability analysis or an equivalent metric. & Held-out data, a metric suited to label prevalence, and results before and after any recalibration are preferable. Raw probabilities are miscalibrated in some workloads; recalibration reduces error by 3.3$\times$, while a 0.01 grid limits rare-label resolution~\citep{rafe2026crash}. & 13/17; 1 unclear \\
C2 & Accuracy among the highest-confidence examples, by task. & The corpus contains high confidence at near-base-rate accuracy~\citep{ibrahim2026css} and wrong tool calls with mean confidence .778~\citep{wu2026reflex}; under optimized answer-preserving context, 45\% of Jev's initially correct decisions (73\% of those redirected) end with probability $\geq$0.7 on the wrong option~\citep{xu2026jevout}. & -- \\
C3 & Where confidence thresholds gate acceptance or escalation: thresholds fitted on data separate from the evaluation set, or fixed before evaluation and not tuned on it. & Frozen thresholds should also be tested for transfer across workloads or fallback models, since thresholds fitted on a selection set did not hold on held-out items for every fallback, and fitted temperatures differ across workloads~\citep{li2026judge}; some fixed thresholds are applied to scores the authors explicitly do not assume to be calibrated~\citep{jiang2026jevmem}. & 7/7; 3 unclear \\
C4 & Where TDM latency is reported or a latency advantage is claimed: latency both with and without caching. & Serving location, concurrency and measured cost per correct outcome should also be stated. Caching removes one reported latency advantage~\citep{li2026edge}; two sibling papers report opposite cost directions on different workloads~\citep{li2026edge,li2026sixg}. & 3/20; 1 unclear \\
\bottomrule
\end{tabularx}
\caption{Evaluation checklist for TDMs. \emph{Count}: compliant papers over papers to which the item applies, among the 27 papers that evaluate a model (criteria in Appendix~\ref{app:coding}); papers coded as unclear are excluded from both counts; ``--'' marks disclosures not reliably scorable from the text.}
\label{tab:checklist}
\end{table}

The checklist also implies a minimal reporting card for future TDM results: exact model version and run date; test difficulty or model spread; one ordinary-LM readout baseline; one cheap generative baseline; intervals from repeated runs; held-out calibration; and, when a cascade is used, a frozen threshold evaluated on a new workload or fallback model.

\section{Open Problems}
\label{sec:open}

\paragraph{Attributing Jev's gains.}
Direct comparisons between Jev and open systems do not produce a consistent ordering. Jev leads the best open decision system on social-science annotation~\citep{ibrahim2026css}, while an open 2B model outperforms it on the grid-world decisions for which that model was trained~\citep{cheng2026thisthat}. A useful ablation would compare matched-cost open backbones under three conditions: no task training, task fine-tuning with ordinary label probabilities, and calibration-oriented decision training.

\paragraph{Decisions that need computation.}
The current interface provides no explicit iterative computation or generated intermediate reasoning. The corpus suggests that this limitation matters on multi-step arithmetic and derivation checking~\citep{cheng2026thisthat,li2026judge}, while the one long-horizon study, a single game confounded by an affordability filter, pairs the TDM with an external planner whose contribution was not isolated~\citep{ma2026jevstar}. Decomposing a hard decision into atomic questions may help~\citep{typesafe2026docs}, but the resulting error propagation and cost trade-off have not been measured systematically.

\paragraph{Faithfulness to option semantics.}
\citet{sun2026optionname} show that a TDM can follow an option name rather than the rubric bound to it, and answer-preserving context or a change of primitive can move decisions as well~\citep{rao2026rubricjudge,xu2026jevout}. Future work should test invariance to option renaming and ordering and distinguish genuine rubric following from surface matching.

\paragraph{Calibration under deployment shift.}
A useful confidence signal should remain meaningful when prevalence, workload, fallback model, or hosted-model version changes. Existing studies show failures along several of these axes~\citep{ibrahim2026css,li2026judge,rafe2026crash}. Longitudinal evaluation across Jev versions would be particularly valuable because all current hosted studies cover essentially the same release window.

\paragraph{When does a TDM cascade beat a generative cascade?}
Confidence-based deferral is a recurring pattern in the corpus, but the one study that compares against a cheap generative cascade finds the TDM cascade ahead only on its own benchmark; on the external $\tau^2$-bench the generative cascade is cheaper with no significant accuracy difference~\citep{wu2026reflex}. Correlated errors between stages can also cap what escalation recovers~\citep{rao2026rubricjudge}. Relevant variables likely include option cardinality, the number of questions sharing a state, latency budget, the share of easy examples, and fallback cost. A shared, non-saturated benchmark that varies these conditions would make the comparison more informative.

\section{Limitations}
\label{sec:limitations}

This review covers a small and unusually young corpus: 28 papers written within about nine days of a product release, most of them evaluating the same hosted version. New papers appeared at several per day while we worked, so the corpus is a snapshot as of 25 September 2026. Many are preprints and may change. Our search focused on arXiv and used blogs, documentation, and model cards only for vendor facts, so industrial experience is underrepresented. The keyword-based search may miss systems that expose similar interfaces without using Jev or typed-decision terminology. Several checklist items require judgment, and we leave items unscored where the text does not support a reliable decision. Finally, because Jev's internals are closed, conclusions about mechanism come from open TDMs and need not describe the hosted model.

\section{Conclusion}
\label{sec:conclusion}

Typed decision models make a familiar idea --- scoring a small set of labels --- available as a software-facing decision interface with caller-defined options, guaranteed output support, shared processing across multiple questions, and probabilities intended for routing. The first week of evidence supports clear efficiency gains in some workloads, but not a general accuracy advantage for the typed readout itself. Calibration remains workload-dependent, and several of the strongest deployment results use the TDM as an inexpensive first stage of a cascade rather than as a universal replacement for generative models. The evaluation checklist is intended to make the next round of evidence easier to compare, reproduce, and trust.

\section*{Use of AI Tools}

LLM-based tools were used to assist with literature organization, data extraction, reference verification, and manuscript preparation. All quantitative results and references were checked against the original sources. The authors reviewed and edited the manuscript and take full responsibility for its content and conclusions.

\bibliographystyle{plainnat}
\bibliography{jev,refs}

\begin{thebibliography}{87}
\providecommand{\natexlab}[1]{#1}
\providecommand{\url}[1]{\texttt{#1}}
\expandafter\ifx\csname urlstyle\endcsname\relax
  \providecommand{\doi}[1]{doi: #1}\else
  \providecommand{\doi}{doi: \begingroup \urlstyle{rm}\Url}\fi

\bibitem[Abbasi et~al.(2026)Abbasi, Aqrawi, and Kwartler]{abbasi2026harness}
Arian Abbasi, Alan Aqrawi, and Ted Kwartler.
\newblock {Control the Harness, Control the Cost: Routing and Governing AI
  Coding Agents in the Enterprise}.
\newblock arXiv preprint arXiv:2609.28919, 2026.

\bibitem[Aggarwal et~al.(2024)Aggarwal, Madaan, Anand, Potharaju, Mishra, Zhou,
  Gupta, Rajagopal, Kappaganthu, Yang, Upadhyay, Faruqui, and
  Mausam]{aggarwal2024automix}
Pranjal Aggarwal, Aman Madaan, Ankit Anand, Srividya~Pranavi Potharaju, Swaroop
  Mishra, Pei Zhou, Aditya Gupta, Dheeraj Rajagopal, Karthik Kappaganthu,
  Yiming Yang, Shyam Upadhyay, Manaal Faruqui, and Mausam.
\newblock {AutoMix: Automatically Mixing Language Models}.
\newblock In \emph{Advances in Neural Information Processing Systems}, 2024.

\bibitem[Almeida(2026)]{typesafe2026jev}
Diogo Almeida.
\newblock {Introducing System One Models \& Jev}.
\newblock TypeSafe AI blog,
  \url{https://typesafe.ai/blog/introducing-system-one-models-and-jev},
  September 2026.
\newblock Accessed 2026-09-24.

\bibitem[{Anonymous Authors}(2026)]{anon2026decisions}
{Anonymous Authors}.
\newblock {Decisions, Not Tokens: A Survey of Typed, Calibrated, and
  Machine-Native Decision Models from Classical Classifiers to Jev}.
\newblock \url{https://github.com/youzizzz1028/Awesome-Jev}, 2026.
\newblock Full working draft, 21 September 2026. Accessed 2026-09-25.

\bibitem[Banerjee et~al.(2025)Banerjee, Suresh, Ugare, Misailovic, and
  Singh]{banerjee2025crane}
Debangshu Banerjee, Tarun Suresh, Shubham Ugare, Sasa Misailovic, and Gagandeep
  Singh.
\newblock {CRANE: Reasoning with constrained LLM generation}.
\newblock In \emph{International Conference on Machine Learning (ICML)}, 2025.

\bibitem[Bani-Harouni et~al.(2026)Bani-Harouni, Pellegrini, Stangel, Özsoy,
  Zaripova, Navab, and Keicher]{baniharouni2026doubt}
David Bani-Harouni, Chantal Pellegrini, Paul Stangel, Ege Özsoy, Kamilia
  Zaripova, Nassir Navab, and Matthias Keicher.
\newblock {Rewarding Doubt: A Reinforcement Learning Approach to Calibrated
  Confidence Expression of Large Language Models}.
\newblock In \emph{International Conference on Learning Representations
  (ICLR)}, 2026.

\bibitem[Baumann et~al.(2025)Baumann, Röttger, Urman, Wendsjö,
  {Plaza-del-Arco}, Gruber, and Hovy]{baumann2025hacking}
Joachim Baumann, Paul Röttger, Aleksandra Urman, Albert Wendsjö, Flor~Miriam
  {Plaza-del-Arco}, Johannes~B. Gruber, and Dirk Hovy.
\newblock {Large Language Model Hacking: Quantifying the Hidden Risks of Using
  LLMs for Text Annotation}.
\newblock arXiv preprint arXiv:2509.08825, 2025.

\bibitem[Beurer-Kellner et~al.(2024)Beurer-Kellner, Fischer, and
  Vechev]{beurerkellner2024domino}
Luca Beurer-Kellner, Marc Fischer, and Martin Vechev.
\newblock {Guiding LLMs The Right Way: Fast, Non-Invasive Constrained
  Generation}.
\newblock In \emph{International Conference on Machine Learning (ICML)}, pages
  3658--3673, 2024.

\bibitem[Bradley and Terry(1952)]{bradley1952rank}
Ralph~Allan Bradley and Milton~E. Terry.
\newblock {Rank Analysis of Incomplete Block Designs: I. The Method of Paired
  Comparisons}.
\newblock \emph{Biometrika}, 39\penalty0 (3/4):\penalty0 324--345, 1952.
\newblock \doi{10.2307/2334029}.

\bibitem[Brown et~al.(2020)Brown, Mann, Ryder, Subbiah, Kaplan, Dhariwal,
  Neelakantan, Shyam, Sastry, Askell, Agarwal, Herbert-Voss, Krueger, Henighan,
  Child, Ramesh, Ziegler, Wu, Winter, Hesse, Chen, Sigler, Litwin, Gray, Chess,
  Clark, Berner, McCandlish, Radford, Sutskever, and Amodei]{brown2020gpt3}
Tom~B. Brown, Benjamin Mann, Nick Ryder, Melanie Subbiah, Jared Kaplan,
  Prafulla Dhariwal, Arvind Neelakantan, Pranav Shyam, Girish Sastry, Amanda
  Askell, Sandhini Agarwal, Ariel Herbert-Voss, Gretchen Krueger, Tom Henighan,
  Rewon Child, Aditya Ramesh, Daniel~M. Ziegler, Jeffrey Wu, Clemens Winter,
  Christopher Hesse, Mark Chen, Eric Sigler, Mateusz Litwin, Scott Gray,
  Benjamin Chess, Jack Clark, Christopher Berner, Sam McCandlish, Alec Radford,
  Ilya Sutskever, and Dario Amodei.
\newblock {Language Models are Few-Shot Learners}.
\newblock In \emph{Advances in Neural Information Processing Systems}, 2020.

\bibitem[Chen et~al.(2024)Chen, Zaharia, and Zou]{chen2024frugalgpt}
Lingjiao Chen, Matei Zaharia, and James Zou.
\newblock {FrugalGPT: How to Use Large Language Models While Reducing Cost and
  Improving Performance}.
\newblock \emph{Transactions on Machine Learning Research}, 2024.

\bibitem[Cheng et~al.(2026)Cheng, Dai, and Sun]{cheng2026thisthat}
Zehua Cheng, Wei Dai, and Jiahao Sun.
\newblock {this-that-model-1.0: A typed decision model that decides in 30 ms,
  for a millionth of a cent}.
\newblock arXiv preprint arXiv:2609.23886, 2026.

\bibitem[Chow(1970)]{chow1970reject}
C.-K. Chow.
\newblock {On Optimum Recognition Error and Reject Tradeoff}.
\newblock \emph{IEEE Transactions on Information Theory}, 16\penalty0
  (1):\penalty0 41--46, 1970.
\newblock \doi{10.1109/TIT.1970.1054406}.

\bibitem[Da{\u{g}}l{\i} et~al.(2026)Da{\u{g}}l{\i}, Da{\u{g}}l{\i}, and
  Da{\u{g}}l{\i}]{dagli2026fractal}
Volkan Da{\u{g}}l{\i}, Zerrin Da{\u{g}}l{\i}, and Da{\u{g}}han Da{\u{g}}l{\i}.
\newblock {Universal Fractal Natural Language Decision Map: Real-Time Edge
  Triage Across Heterogeneous Domains}.
\newblock arXiv preprint arXiv:2609.25498, 2026.

\bibitem[Damani et~al.(2026)Damani, Puri, Slocum, Shenfeld, Choshen, Kim, and
  Andreas]{damani2026rlcr}
Mehul Damani, Isha Puri, Stewart Slocum, Idan Shenfeld, Leshem Choshen, Yoon
  Kim, and Jacob Andreas.
\newblock {Beyond Binary Rewards: Training LMs to Reason About Their
  Uncertainty}.
\newblock In \emph{International Conference on Learning Representations
  (ICLR)}, 2026.

\bibitem[Deng et~al.(2026)Deng, Fan, Zhang, and Xie]{deng2026science}
Boyuan Deng, Shuyi Fan, Hongyang Zhang, and Xinhong Xie.
\newblock {Jev for Scientific Decisions: Evaluating Semantic Choices and Their
  Consequences}.
\newblock arXiv preprint arXiv:2609.24965, 2026.
\newblock Version 2, 23 September 2026.

\bibitem[Dong et~al.(2025)Dong, Ruan, Cai, Xu, Zhao, Lai, and
  Chen]{dong2025xgrammar}
Yixin Dong, Charlie~F. Ruan, Yaxing Cai, Ziyi Xu, Yilong Zhao, Ruihang Lai, and
  Tianqi Chen.
\newblock {XGrammar: Flexible and Efficient Structured Generation Engine for
  Large Language Models}.
\newblock In \emph{Proceedings of Machine Learning and Systems (MLSys)}, 2025.

\bibitem[dos Santos~Barbosa(2026)]{barbosa2026pentest}
Joas~Antonio dos Santos~Barbosa.
\newblock {Calibrated Decision Models for Autonomous Penetration-Testing
  Harnesses: JEV and Laya as System One Decision Layers for LLM-Driven Pentest
  Agents}.
\newblock arXiv preprint arXiv:2609.28940, 2026.

\bibitem[Egami et~al.(2023)Egami, Hinck, Stewart, and Wei]{egami2023dsl}
Naoki Egami, Musashi Hinck, Brandon~M. Stewart, and Hanying Wei.
\newblock {Using Imperfect Surrogates for Downstream Inference: Design-based
  Supervised Learning for Social Science Applications of Large Language
  Models}.
\newblock In \emph{Advances in Neural Information Processing Systems},
  volume~36, 2023.

\bibitem[Gao et~al.(2021)Gao, Fisch, and Chen]{gao2021lmbff}
Tianyu Gao, Adam Fisch, and Danqi Chen.
\newblock Making pre-trained language models better few-shot learners.
\newblock In Chengqing Zong, Fei Xia, Wenjie Li, and Roberto Navigli, editors,
  \emph{Proceedings of the 59th Annual Meeting of the Association for
  Computational Linguistics and the 11th International Joint Conference on
  Natural Language Processing (Volume 1: Long Papers)}, pages 3816--3830,
  Online, August 2021. Association for Computational Linguistics.
\newblock \doi{10.18653/v1/2021.acl-long.295}.
\newblock URL \url{https://aclanthology.org/2021.acl-long.295/}.

\bibitem[Geifman and El-Yaniv(2017)]{geifman2017selective}
Yonatan Geifman and Ran El-Yaniv.
\newblock {Selective Classification for Deep Neural Networks}.
\newblock In \emph{Advances in Neural Information Processing Systems}, 2017.

\bibitem[Geng et~al.(2023)Geng, Josifoski, Peyrard, and West]{geng2023gcd}
Saibo Geng, Martin Josifoski, Maxime Peyrard, and Robert West.
\newblock Grammar-constrained decoding for structured {NLP} tasks without
  finetuning.
\newblock In Houda Bouamor, Juan Pino, and Kalika Bali, editors,
  \emph{Proceedings of the 2023 Conference on Empirical Methods in Natural
  Language Processing}, pages 10932--10952, Singapore, December 2023.
  Association for Computational Linguistics.
\newblock \doi{10.18653/v1/2023.emnlp-main.674}.
\newblock URL \url{https://aclanthology.org/2023.emnlp-main.674/}.

\bibitem[Guo et~al.(2017)Guo, Pleiss, Sun, and Weinberger]{guo2017calibration}
Chuan Guo, Geoff Pleiss, Yu~Sun, and Kilian~Q. Weinberger.
\newblock {On Calibration of Modern Neural Networks}.
\newblock In \emph{International Conference on Machine Learning (ICML)}, 2017.

\bibitem[Guo et~al.(2026)Guo, Liu, Deng, Li, Zhao, Wu, Chen, Zhang, and
  Zhang]{guo2026justaskjev}
Ruoqi Guo, Yi~Liu, Gelei Deng, Yuekang Li, Lida Zhao, Yutao Wu, Simin Chen,
  Ying Zhang, and Leo~Yu Zhang.
\newblock {Just Ask Jev: Reinforcement Learning for Calibrated Decisions as a
  Zero-Shot Detector of AI Alignment Failures}.
\newblock arXiv preprint arXiv:2609.29429, 2026.

\bibitem[Gupta et~al.(2024)Gupta, Narasimhan, Jitkrittum, Rawat, Menon, and
  Kumar]{gupta2024cascades}
Neha Gupta, Harikrishna Narasimhan, Wittawat Jitkrittum, Ankit~Singh Rawat,
  Aditya~Krishna Menon, and Sanjiv Kumar.
\newblock {Language Model Cascades: Token-level uncertainty and beyond}.
\newblock In \emph{International Conference on Learning Representations
  (ICLR)}, 2024.

\bibitem[Holtzman et~al.(2021)Holtzman, West, Shwartz, Choi, and
  Zettlemoyer]{holtzman2021surface}
Ari Holtzman, Peter West, Vered Shwartz, Yejin Choi, and Luke Zettlemoyer.
\newblock Surface form competition: Why the highest probability answer isn{'}t
  always right.
\newblock In Marie-Francine Moens, Xuanjing Huang, Lucia Specia, and Scott
  Wen-tau Yih, editors, \emph{Proceedings of the 2021 Conference on Empirical
  Methods in Natural Language Processing}, pages 7038--7051, Online and Punta
  Cana, Dominican Republic, November 2021. Association for Computational
  Linguistics.
\newblock \doi{10.18653/v1/2021.emnlp-main.564}.
\newblock URL \url{https://aclanthology.org/2021.emnlp-main.564/}.

\bibitem[Huang et~al.(2026)Huang, Yang, Ma, Liang, Cai, Ma, Chen, Sun, and
  Tan]{huang2026radiology}
Jiaju Huang, Hao Yang, Xinyu Ma, Xinglong Liang, Kunyan Cai, Junqiang Ma,
  Shaobin Chen, Yue Sun, and Tao Tan.
\newblock {Can Jev Judge Radiology Reports? Evaluating a System One Model for
  Clinical Factuality}.
\newblock arXiv preprint arXiv:2609.27607, 2026.

\bibitem[Ibrahim and Zaki(2026)]{ibrahim2026css}
Hazem Ibrahim and Yasir Zaki.
\newblock {Evaluating Decision Models for Text Annotation in Computational
  Social Science}.
\newblock arXiv preprint arXiv:2609.24574, 2026.
\newblock Version 2, 23 September 2026.

\bibitem[Jiang et~al.(2026)Jiang, Li, and Li]{jiang2026jevmem}
Dongming Jiang, Yi~Li, and Bingzhe Li.
\newblock {Jev-Mem: System-One-Controlled Agentic Memory for Efficient AI
  Agents}.
\newblock arXiv preprint arXiv:2609.23986, 2026.

\bibitem[Jitkrittum et~al.(2023)Jitkrittum, Gupta, Menon, Narasimhan, Rawat,
  and Kumar]{jitkrittum2023deferral}
Wittawat Jitkrittum, Neha Gupta, Aditya~Krishna Menon, Harikrishna Narasimhan,
  Ankit~Singh Rawat, and Sanjiv Kumar.
\newblock {When Does Confidence-Based Cascade Deferral Suffice?}
\newblock In \emph{Advances in Neural Information Processing Systems}, 2023.

\bibitem[Kadavath et~al.(2022)Kadavath, Conerly, Askell, Henighan, Drain,
  Perez, Schiefer, Hatfield-Dodds, DasSarma, Tran-Johnson, Johnston, El-Showk,
  Jones, Elhage, Hume, Chen, Bai, Bowman, Fort, Ganguli, Hernandez, Jacobson,
  Kernion, Kravec, Lovitt, Ndousse, Olsson, Ringer, Amodei, Brown, Clark,
  Joseph, Mann, McCandlish, Olah, and Kaplan]{kadavath2022know}
Saurav Kadavath, Tom Conerly, Amanda Askell, Tom Henighan, Dawn Drain, Ethan
  Perez, Nicholas Schiefer, Zac Hatfield-Dodds, Nova DasSarma, Eli
  Tran-Johnson, Scott Johnston, Sheer El-Showk, Andy Jones, Nelson Elhage,
  Tristan Hume, Anna Chen, Yuntao Bai, Sam Bowman, Stanislav Fort, Deep
  Ganguli, Danny Hernandez, Josh Jacobson, Jackson Kernion, Shauna Kravec,
  Liane Lovitt, Kamal Ndousse, Catherine Olsson, Sam Ringer, Dario Amodei, Tom
  Brown, Jack Clark, Nicholas Joseph, Ben Mann, Sam McCandlish, Chris Olah, and
  Jared Kaplan.
\newblock {Language Models (Mostly) Know What They Know}.
\newblock arXiv preprint arXiv:2207.05221, 2022.

\bibitem[Kahneman(2011)]{kahneman2011thinking}
Daniel Kahneman.
\newblock \emph{{Thinking, Fast and Slow}}.
\newblock Farrar, Straus and Giroux, New York, 2011.

\bibitem[Kamath et~al.(2020)Kamath, Jia, and Liang]{kamath2020selective}
Amita Kamath, Robin Jia, and Percy Liang.
\newblock Selective question answering under domain shift.
\newblock In Dan Jurafsky, Joyce Chai, Natalie Schluter, and Joel Tetreault,
  editors, \emph{Proceedings of the 58th Annual Meeting of the Association for
  Computational Linguistics}, pages 5684--5696, Online, July 2020. Association
  for Computational Linguistics.
\newblock \doi{10.18653/v1/2020.acl-main.503}.
\newblock URL \url{https://aclanthology.org/2020.acl-main.503/}.

\bibitem[Kim et~al.(2024)Kim, Shin, Cho, Jang, Longpre, Lee, Yun, Shin, Kim,
  Thorne, and Seo]{kim2024prometheus}
Seungone Kim, Jamin Shin, Yejin Cho, Joel Jang, Shayne Longpre, Hwaran Lee,
  Sangdoo Yun, Seongjin Shin, Sungdong Kim, James Thorne, and Minjoon Seo.
\newblock {Prometheus: Inducing Fine-grained Evaluation Capability in Language
  Models}.
\newblock In \emph{International Conference on Learning Representations
  (ICLR)}, 2024.

\bibitem[Laurer et~al.(2023)Laurer, van Atteveldt, Casas, and
  Welbers]{laurer2023universal}
Moritz Laurer, Wouter van Atteveldt, Andreu Casas, and Kasper Welbers.
\newblock {Building Efficient Universal Classifiers with Natural Language
  Inference}.
\newblock arXiv preprint arXiv:2312.17543, 2023.

\bibitem[Li et~al.(2026{\natexlab{a}})Li, Wang, Gong, Lang, and Yu]{li2026edge}
Delong Li, Xu~Wang, Haochen Gong, Rui Lang, and Guangsheng Yu.
\newblock {Replacing Large Language Models with Jev Decision Models for
  Low-Latency Edge Service Orchestration}.
\newblock arXiv preprint arXiv:2609.22753, 2026{\natexlab{a}}.

\bibitem[Li et~al.(2026{\natexlab{b}})Li, Wang, Gong, Lang, and Yu]{li2026sixg}
Delong Li, Xu~Wang, Haochen Gong, Rui Lang, and Guangsheng Yu.
\newblock {Fast Intent-Driven Service Orchestration with Jev for 6G Edge
  Networks}.
\newblock arXiv preprint arXiv:2609.23136, 2026{\natexlab{b}}.

\bibitem[Li(2026)]{li2026kite}
Hengyu Li.
\newblock {KITE: Scaling Jev Population Experiments with Sparse Flagship
  Calibration}.
\newblock arXiv preprint arXiv:2609.27535, 2026.

\bibitem[Li et~al.(2026{\natexlab{c}})Li, Miao, Krishnan, and
  Padman]{li2026judge}
Yubo Li, Yidi Miao, Ramayya Krishnan, and Rema Padman.
\newblock {JEV-as-a-Judge: Accept When Confident, Escalate When Unsure}.
\newblock arXiv preprint arXiv:2609.26550, 2026{\natexlab{c}}.

\bibitem[Ling et~al.(2026)Ling, Xue, and Ye]{ling2026wild}
Guoming Ling, Muen Xue, and Zijian Ye.
\newblock {Jev in the Wild: A Data-Driven Analysis of the Jev Model's
  Functionality, Applications and Ecosystem}.
\newblock arXiv preprint arXiv:2609.30216, 2026.

\bibitem[Liu et~al.(2023)Liu, Iter, Xu, Wang, Xu, and Zhu]{liu2023geval}
Yang Liu, Dan Iter, Yichong Xu, Shuohang Wang, Ruochen Xu, and Chenguang Zhu.
\newblock {G}-eval: {NLG} evaluation using gpt-4 with better human alignment.
\newblock In Houda Bouamor, Juan Pino, and Kalika Bali, editors,
  \emph{Proceedings of the 2023 Conference on Empirical Methods in Natural
  Language Processing}, pages 2511--2522, Singapore, December 2023. Association
  for Computational Linguistics.
\newblock \doi{10.18653/v1/2023.emnlp-main.153}.
\newblock URL \url{https://aclanthology.org/2023.emnlp-main.153/}.

\bibitem[Liu et~al.(2025)Liu, Yao, Min, Cao, Hou, and Li]{liu2025rmbench}
Yantao Liu, Zijun Yao, Rui Min, Yixin Cao, Lei Hou, and Juanzi Li.
\newblock {RM-Bench: Benchmarking Reward Models of Language Models with
  Subtlety and Style}.
\newblock In \emph{International Conference on Learning Representations
  (ICLR)}, 2025.

\bibitem[Luo(2026)]{luo2026jevsurvey}
Meng Luo.
\newblock {Jev and Typed Decision Models: An Empirical Survey of Calibration,
  Selective Control, and Open Implementations}.
\newblock \url{https://github.com/Eurekaleo/awesome-jev-survey}, 2026.
\newblock Working draft, version 0.2.0, data cutoff 24 September 2026. Accessed
  2026-09-25.

\bibitem[Ma et~al.(2026)Ma, Zhao, Zeng, and Zhao]{ma2026jevstar}
Weiyu Ma, Liangbing Zhao, Yongcheng Zeng, and Jian Zhao.
\newblock {JEV-Star: Fast, Low-Cost StarCraft II Control with Language-Model
  Planning}.
\newblock arXiv preprint arXiv:2609.27331, 2026.

\bibitem[Madras et~al.(2018)Madras, Pitassi, and Zemel]{madras2018defer}
David Madras, Toniann Pitassi, and Richard Zemel.
\newblock {Predict Responsibly: Improving Fairness and Accuracy by Learning to
  Defer}.
\newblock In \emph{Advances in Neural Information Processing Systems}, pages
  6150--6160, 2018.

\bibitem[Min et~al.(2022)Min, Lewis, Hajishirzi, and Zettlemoyer]{min2022noisy}
Sewon Min, Mike Lewis, Hannaneh Hajishirzi, and Luke Zettlemoyer.
\newblock Noisy channel language model prompting for few-shot text
  classification.
\newblock In Smaranda Muresan, Preslav Nakov, and Aline Villavicencio, editors,
  \emph{Proceedings of the 60th Annual Meeting of the Association for
  Computational Linguistics (Volume 1: Long Papers)}, pages 5316--5330, Dublin,
  Ireland, May 2022. Association for Computational Linguistics.
\newblock \doi{10.18653/v1/2022.acl-long.365}.
\newblock URL \url{https://aclanthology.org/2022.acl-long.365/}.

\bibitem[Mozannar and Sontag(2020)]{mozannar2020defer}
Hussein Mozannar and David Sontag.
\newblock {Consistent Estimators for Learning to Defer to an Expert}.
\newblock In \emph{International Conference on Machine Learning (ICML)}, pages
  7076--7087, 2020.

\bibitem[Nogueira et~al.(2020)Nogueira, Jiang, Pradeep, and
  Lin]{nogueira2020monot5}
Rodrigo Nogueira, Zhiying Jiang, Ronak Pradeep, and Jimmy Lin.
\newblock Document ranking with a pretrained sequence-to-sequence model.
\newblock In Trevor Cohn, Yulan He, and Yang Liu, editors, \emph{Findings of
  the Association for Computational Linguistics: EMNLP 2020}, pages 708--718,
  Online, November 2020. Association for Computational Linguistics.
\newblock \doi{10.18653/v1/2020.findings-emnlp.63}.
\newblock URL \url{https://aclanthology.org/2020.findings-emnlp.63/}.

\bibitem[Ouyang et~al.(2022)Ouyang, Wu, Jiang, Almeida, Wainwright, Mishkin,
  Zhang, Agarwal, Slama, Ray, Schulman, Hilton, Kelton, Miller, Simens, Askell,
  Welinder, Christiano, Leike, and Lowe]{ouyang2022instructgpt}
Long Ouyang, Jeff Wu, Xu~Jiang, Diogo Almeida, Carroll~L. Wainwright, Pamela
  Mishkin, Chong Zhang, Sandhini Agarwal, Katarina Slama, Alex Ray, John
  Schulman, Jacob Hilton, Fraser Kelton, Luke Miller, Maddie Simens, Amanda
  Askell, Peter Welinder, Paul Christiano, Jan Leike, and Ryan Lowe.
\newblock {Training language models to follow instructions with human
  feedback}.
\newblock In \emph{Advances in Neural Information Processing Systems}, 2022.

\bibitem[Park et~al.(2024)Park, Wang, Berg-Kirkpatrick, Polikarpova, and
  D'Antoni]{park2024gad}
Kanghee Park, Jiayu Wang, Taylor Berg-Kirkpatrick, Nadia Polikarpova, and Loris
  D'Antoni.
\newblock {Grammar-Aligned Decoding}.
\newblock In \emph{Advances in Neural Information Processing Systems}, 2024.

\bibitem[Pezeshkpour and Hruschka(2024)]{pezeshkpour2024order}
Pouya Pezeshkpour and Estevam Hruschka.
\newblock Large language models sensitivity to the order of options in
  multiple-choice questions.
\newblock In Kevin Duh, Helena Gomez, and Steven Bethard, editors,
  \emph{Findings of the Association for Computational Linguistics: NAACL 2024},
  pages 2006--2017, Mexico City, Mexico, June 2024. Association for
  Computational Linguistics.
\newblock \doi{10.18653/v1/2024.findings-naacl.130}.
\newblock URL \url{https://aclanthology.org/2024.findings-naacl.130/}.

\bibitem[Rafe and Das(2026)]{rafe2026crash}
Amir Rafe and Subasish Das.
\newblock {Calibrated Decisions at Scale: Converting Police Crash Narratives
  into Probabilistic Crash Variables with a System One Model (Jev)}.
\newblock arXiv preprint arXiv:2609.24052, 2026.

\bibitem[Rao and Callison-Burch(2026)]{rao2026rubricjudge}
Delip Rao and Chris Callison-Burch.
\newblock {JEV vs. LLMs as Rubric Judges: Cheaper, Faster, and Wrong in the
  Same Places}.
\newblock arXiv preprint arXiv:2609.29769, 2026.

\bibitem[Ren et~al.(2026)Ren, Zewde, Shen, Zhou, Ng, Raj, Duong, Zhang, and
  Tiangratanakul]{ren2026callscreen}
Simiao Ren, Kidus Zewde, Xingyu Shen, Yuchen Zhou, Dennis Ng, Ankit Raj, Tommy
  Duong, Yuxin Zhang, and Neo Tiangratanakul.
\newblock {Open-Jev Judgments on CallScreenBench: Calibrated One-Pass Scam
  Screening with a Small Language Model}.
\newblock arXiv preprint arXiv:2609.23959, 2026.

\bibitem[Robinson and Wingate(2023)]{robinson2023mcqa}
Joshua Robinson and David Wingate.
\newblock {Leveraging Large Language Models for Multiple Choice Question
  Answering}.
\newblock In \emph{International Conference on Learning Representations
  (ICLR)}, 2023.

\bibitem[Robitza(2026)]{robitza2026jevqa}
Werner Robitza.
\newblock {JEVQA - Video Quality from Metadata, Bitstream, and Pixel Features
  with a General-Purpose Decision Model}.
\newblock arXiv preprint arXiv:2609.24395, 2026.

\bibitem[Schick and Sch{\"u}tze(2021{\natexlab{a}})]{schick2021pet}
Timo Schick and Hinrich Sch{\"u}tze.
\newblock Exploiting cloze-questions for few-shot text classification and
  natural language inference.
\newblock In Paola Merlo, Jorg Tiedemann, and Reut Tsarfaty, editors,
  \emph{Proceedings of the 16th Conference of the European Chapter of the
  Association for Computational Linguistics: Main Volume}, pages 255--269,
  Online, April 2021{\natexlab{a}}. Association for Computational Linguistics.
\newblock \doi{10.18653/v1/2021.eacl-main.20}.
\newblock URL \url{https://aclanthology.org/2021.eacl-main.20/}.

\bibitem[Schick and Sch{\"u}tze(2021{\natexlab{b}})]{schick2021small}
Timo Schick and Hinrich Sch{\"u}tze.
\newblock It{'}s not just size that matters: Small language models are also
  few-shot learners.
\newblock In Kristina Toutanova, Anna Rumshisky, Luke Zettlemoyer, Dilek
  Hakkani-Tur, Iz~Beltagy, Steven Bethard, Ryan Cotterell, Tanmoy Chakraborty,
  and Yichao Zhou, editors, \emph{Proceedings of the 2021 Conference of the
  North American Chapter of the Association for Computational Linguistics:
  Human Language Technologies}, pages 2339--2352, Online, June
  2021{\natexlab{b}}. Association for Computational Linguistics.
\newblock \doi{10.18653/v1/2021.naacl-main.185}.
\newblock URL \url{https://aclanthology.org/2021.naacl-main.185/}.

\bibitem[Scholak et~al.(2021)Scholak, Schucher, and
  Bahdanau]{scholak2021picard}
Torsten Scholak, Nathan Schucher, and Dzmitry Bahdanau.
\newblock {PICARD}: Parsing incrementally for constrained auto-regressive
  decoding from language models.
\newblock In Marie-Francine Moens, Xuanjing Huang, Lucia Specia, and Scott
  Wen-tau Yih, editors, \emph{Proceedings of the 2021 Conference on Empirical
  Methods in Natural Language Processing}, pages 9895--9901, Online and Punta
  Cana, Dominican Republic, November 2021. Association for Computational
  Linguistics.
\newblock \doi{10.18653/v1/2021.emnlp-main.779}.
\newblock URL \url{https://aclanthology.org/2021.emnlp-main.779/}.

\bibitem[Stepanov et~al.(2025)Stepanov, Shtopko, Vodianytskyi, Lukashov,
  Yavorskyi, and Yaroshenko]{stepanov2025gliclass}
Ihor Stepanov, Mykhailo Shtopko, Dmytro Vodianytskyi, Oleksandr Lukashov,
  Alexander Yavorskyi, and Mykyta Yaroshenko.
\newblock {GLiClass: Generalist Lightweight Model for Sequence Classification
  Tasks}.
\newblock arXiv preprint arXiv:2508.07662, 2025.

\bibitem[Sun et~al.(2026)Sun, Xu, Shi, and Yang]{sun2026optionname}
Yu~Sun, Junhao Xu, Jiajia Shi, and Zijin Yang.
\newblock {Type-Safe Is Not Error-Free: A Constrained Decision Head Follows the
  Option Name, Not the Rubric Bound to It}.
\newblock arXiv preprint arXiv:2609.26758, 2026.
\newblock Version 2, 23 September 2026.

\bibitem[Tam et~al.(2024)Tam, Wu, Tsai, Lin, Lee, and Chen]{tam2024speak}
Zhi~Rui Tam, Cheng-Kuang Wu, Yi-Lin Tsai, Chieh-Yen Lin, Hung-yi Lee, and
  Yun-Nung Chen.
\newblock Let me speak freely? a study on the impact of format restrictions on
  large language model performance.
\newblock In Franck Dernoncourt, Daniel Preo{\c{t}}iuc-Pietro, and Anastasia
  Shimorina, editors, \emph{Proceedings of the 2024 Conference on Empirical
  Methods in Natural Language Processing: Industry Track}, pages 1218--1236,
  Miami, Florida, US, November 2024. Association for Computational Linguistics.
\newblock \doi{10.18653/v1/2024.emnlp-industry.91}.
\newblock URL \url{https://aclanthology.org/2024.emnlp-industry.91/}.

\bibitem[Tan et~al.(2025)Tan, Zhuang, Montgomery, Tang, Cuadron, Wang, Popa,
  and Stoica]{tan2025judgebench}
Sijun Tan, Siyuan Zhuang, Kyle Montgomery, William~Y. Tang, Alejandro Cuadron,
  Chenguang Wang, Raluca~Ada Popa, and Ion Stoica.
\newblock {JudgeBench: A Benchmark for Evaluating LLM-based Judges}.
\newblock In \emph{International Conference on Learning Representations
  (ICLR)}, 2025.

\bibitem[Tian et~al.(2023)Tian, Mitchell, Zhou, Sharma, Rafailov, Yao, Finn,
  and Manning]{tian2023justask}
Katherine Tian, Eric Mitchell, Allan Zhou, Archit Sharma, Rafael Rafailov,
  Huaxiu Yao, Chelsea Finn, and Christopher Manning.
\newblock Just ask for calibration: Strategies for eliciting calibrated
  confidence scores from language models fine-tuned with human feedback.
\newblock In Houda Bouamor, Juan Pino, and Kalika Bali, editors,
  \emph{Proceedings of the 2023 Conference on Empirical Methods in Natural
  Language Processing}, pages 5433--5442, Singapore, December 2023. Association
  for Computational Linguistics.
\newblock \doi{10.18653/v1/2023.emnlp-main.330}.
\newblock URL \url{https://aclanthology.org/2023.emnlp-main.330/}.

\bibitem[{TypeSafe AI}(2026{\natexlab{a}})]{typesafe2026adapter}
{TypeSafe AI}.
\newblock {System One Adapter (Python)}.
\newblock \url{https://github.com/typesafe-ai/system-one-adapter-python},
  2026{\natexlab{a}}.
\newblock Accessed 2026-09-24.

\bibitem[{TypeSafe AI}(2026{\natexlab{b}})]{typesafe2026docs}
{TypeSafe AI}.
\newblock {TypeSafe AI Documentation}.
\newblock \url{https://docs.typesafe.ai/introduction}, 2026{\natexlab{b}}.
\newblock Accessed 2026-09-24.

\bibitem[Varshney and Baral(2022)]{varshney2022cascading}
Neeraj Varshney and Chitta Baral.
\newblock Model cascading: Towards jointly improving efficiency and accuracy of
  {NLP} systems.
\newblock In Yoav Goldberg, Zornitsa Kozareva, and Yue Zhang, editors,
  \emph{Proceedings of the 2022 Conference on Empirical Methods in Natural
  Language Processing}, pages 11007--11021, Abu Dhabi, United Arab Emirates,
  December 2022. Association for Computational Linguistics.
\newblock \doi{10.18653/v1/2022.emnlp-main.756}.
\newblock URL \url{https://aclanthology.org/2022.emnlp-main.756/}.

\bibitem[Wang et~al.(2024)Wang, Ma, Hu, Weber-Genzel, R{\"o}ttger, Kreuter,
  Hovy, and Plank]{wang2024myanswer}
Xinpeng Wang, Bolei Ma, Chengzhi Hu, Leon Weber-Genzel, Paul R{\"o}ttger,
  Frauke Kreuter, Dirk Hovy, and Barbara Plank.
\newblock ``my answer is {C}'': First-token probabilities do not match text
  answers in instruction-tuned language models.
\newblock In Lun-Wei Ku, Andre Martins, and Vivek Srikumar, editors,
  \emph{Findings of the Association for Computational Linguistics: ACL 2024},
  pages 7407--7416, Bangkok, Thailand, August 2024. Association for
  Computational Linguistics.
\newblock \doi{10.18653/v1/2024.findings-acl.441}.
\newblock URL \url{https://aclanthology.org/2024.findings-acl.441/}.

\bibitem[Willard and Louf(2023)]{willard2023outlines}
Brandon~T. Willard and Rémi Louf.
\newblock {Efficient Guided Generation for Large Language Models}.
\newblock arXiv preprint arXiv:2307.09702, 2023.

\bibitem[Wu and Lim(2026{\natexlab{a}})]{wu2026hijacking}
Tiantong Wu and Wei Yang~Bryan Lim.
\newblock {Decision Hijacking: Prompt Injection Attacks on Jev's Typed
  Probabilistic Decisions}.
\newblock arXiv preprint arXiv:2609.28613, 2026{\natexlab{a}}.

\bibitem[Wu and Lim(2026{\natexlab{b}})]{wu2026reflex}
Tiantong Wu and Wei Yang~Bryan Lim.
\newblock {REFLEX with Jev for Efficient Selective Control in LLM Agents}.
\newblock arXiv preprint arXiv:2609.26532, 2026{\natexlab{b}}.

\bibitem[Xiong et~al.(2024)Xiong, Hu, Lu, Li, Fu, He, and
  Hooi]{xiong2024express}
Miao Xiong, Zhiyuan Hu, Xinyang Lu, Yifei Li, Jie Fu, Junxian He, and Bryan
  Hooi.
\newblock {Can LLMs Express Their Uncertainty? An Empirical Evaluation of
  Confidence Elicitation in LLMs}.
\newblock In \emph{International Conference on Learning Representations
  (ICLR)}, 2024.

\bibitem[Xu(2026)]{xu2026jevout}
Zixiang Xu.
\newblock {JevOut: Natural Context Can Flip Decision Models}.
\newblock arXiv preprint arXiv:2609.30243, 2026.

\bibitem[Ye et~al.(2026)Ye, Liu, and Jiang]{ye2026numericjev}
Weiwei Ye, Hangchen Liu, and Renhe Jiang.
\newblock {NumericJev: Jev-like LLM Numerical Decoding with Multiway Decision
  Trees}.
\newblock arXiv preprint arXiv:2609.28587, 2026.

\bibitem[Yin et~al.(2019)Yin, Hay, and Roth]{yin2019benchmarking}
Wenpeng Yin, Jamaal Hay, and Dan Roth.
\newblock Benchmarking zero-shot text classification: Datasets, evaluation and
  entailment approach.
\newblock In Kentaro Inui, Jing Jiang, Vincent Ng, and Xiaojun Wan, editors,
  \emph{Proceedings of the 2019 Conference on Empirical Methods in Natural
  Language Processing and the 9th International Joint Conference on Natural
  Language Processing (EMNLP-IJCNLP)}, pages 3914--3923, Hong Kong, China,
  November 2019. Association for Computational Linguistics.
\newblock \doi{10.18653/v1/D19-1404}.
\newblock URL \url{https://aclanthology.org/D19-1404/}.

\bibitem[Yu and Yao(2026)]{yu2026visualjev}
Guanxu Yu and Yuhang Yao.
\newblock {Visual Jev: Accurate and Efficient Decisions from Shared Visual
  Context}.
\newblock arXiv preprint arXiv:2609.25845, 2026.

\bibitem[Zadrozny and Elkan(2002)]{zadrozny2002transforming}
Bianca Zadrozny and Charles Elkan.
\newblock {Transforming Classifier Scores into Accurate Multiclass Probability
  Estimates}.
\newblock In \emph{Proceedings of the Eighth ACM SIGKDD International
  Conference on Knowledge Discovery and Data Mining}, pages 694--699, 2002.
\newblock \doi{10.1145/775047.775151}.

\bibitem[Zaratiana et~al.(2024)Zaratiana, Tomeh, Holat, and
  Charnois]{zaratiana2024gliner}
Urchade Zaratiana, Nadi Tomeh, Pierre Holat, and Thierry Charnois.
\newblock {GL}i{NER}: Generalist model for named entity recognition using
  bidirectional transformer.
\newblock In Kevin Duh, Helena Gomez, and Steven Bethard, editors,
  \emph{Proceedings of the 2024 Conference of the North American Chapter of the
  Association for Computational Linguistics: Human Language Technologies
  (Volume 1: Long Papers)}, pages 5364--5376, Mexico City, Mexico, June 2024.
  Association for Computational Linguistics.
\newblock \doi{10.18653/v1/2024.naacl-long.300}.
\newblock URL \url{https://aclanthology.org/2024.naacl-long.300/}.

\bibitem[Zhang et~al.(2026)Zhang, Chen, Xie, Zhou, Peng, Fan, Ji, Zheng, Shan,
  Yu, Liu, Chen, Nakov, and He]{zhang2026contract}
Fan Zhang, Yankai Chen, Zhuohan Xie, Yixi Zhou, Sijia Peng, Lei Fan, Xinhua Ji,
  Cunyuan Zheng, Huangyong Shan, Philip~S. Yu, Xue Liu, Yu~Chen, Preslav Nakov,
  and Songwei He.
\newblock {Same Scores, Different Decisions: Evaluating JEV and Language Models
  for Legal Document Understanding}.
\newblock arXiv preprint arXiv:2609.27678, 2026.

\bibitem[Zhang(2026)]{zhang2026jevmobile}
Linghua Zhang.
\newblock {Jev-Mobile: Jev as an Executor for Mobile GUI Agents}.
\newblock arXiv preprint arXiv:2609.30186, 2026.

\bibitem[Zhang et~al.(2025{\natexlab{a}})Zhang, Hosseini, Bansal, Kazemi,
  Kumar, and Agarwal]{zhang2025genverifier}
Lunjun Zhang, Arian Hosseini, Hritik Bansal, Mehran Kazemi, Aviral Kumar, and
  Rishabh Agarwal.
\newblock {Generative Verifiers: Reward Modeling as Next-Token Prediction}.
\newblock In \emph{International Conference on Learning Representations
  (ICLR)}, 2025{\natexlab{a}}.

\bibitem[Zhang et~al.(2025{\natexlab{b}})Zhang, Li, Long, Zhang, Lin, Yang,
  Xie, Yang, Liu, Lin, Huang, and Zhou]{zhang2025qwen3emb}
Yanzhao Zhang, Mingxin Li, Dingkun Long, Xin Zhang, Huan Lin, Baosong Yang,
  Pengjun Xie, An~Yang, Dayiheng Liu, Junyang Lin, Fei Huang, and Jingren Zhou.
\newblock {Qwen3 Embedding: Advancing Text Embedding and Reranking Through
  Foundation Models}.
\newblock arXiv preprint arXiv:2506.05176, 2025{\natexlab{b}}.

\bibitem[Zhao et~al.(2021)Zhao, Wallace, Feng, Klein, and
  Singh]{zhao2021calibrate}
Tony~Z. Zhao, Eric Wallace, Shi Feng, Dan Klein, and Sameer Singh.
\newblock {Calibrate Before Use: Improving Few-Shot Performance of Language
  Models}.
\newblock In \emph{International Conference on Machine Learning (ICML)}, pages
  12697--12706, 2021.

\bibitem[Zheng et~al.(2024)Zheng, Zhou, Meng, Zhou, and Huang]{zheng2024mcq}
Chujie Zheng, Hao Zhou, Fandong Meng, Jie Zhou, and Minlie Huang.
\newblock {Large Language Models Are Not Robust Multiple Choice Selectors}.
\newblock In \emph{International Conference on Learning Representations
  (ICLR)}, 2024.

\bibitem[Zhou et~al.(2024)Zhou, Wan, Proleev, Mincu, Chen, Heller, and
  Roy]{zhou2024batch}
Han Zhou, Xingchen Wan, Lev Proleev, Diana Mincu, Jilin Chen, Katherine Heller,
  and Subhrajit Roy.
\newblock {Batch Calibration: Rethinking Calibration for In-Context Learning
  and Prompt Engineering}.
\newblock In \emph{International Conference on Learning Representations
  (ICLR)}, 2024.

\bibitem[Zhou et~al.(2026)Zhou, Yang, and Zhao]{zhou2026pixeljev}
Xunlan Zhou, Xianliang Yang, and Li~Zhao.
\newblock {From Text Decisions to Pixels: An Study of Jev-Style Visual Choice
  Model}.
\newblock arXiv preprint arXiv:2609.29283, 2026.

\bibitem[Ziems et~al.(2024)Ziems, Held, Shaikh, Chen, Zhang, and
  Yang]{ziems2024css}
Caleb Ziems, William Held, Omar Shaikh, Jiaao Chen, Zhehao Zhang, and Diyi
  Yang.
\newblock Can large language models transform computational social science?
\newblock \emph{Computational Linguistics}, 50\penalty0 (1):\penalty0 237--291,
  March 2024.
\newblock \doi{10.1162/coli_a_00502}.
\newblock URL \url{https://aclanthology.org/2024.cl-1.8/}.

\end{thebibliography}
\appendix
\section{Corpus Versions}
\label{app:versions}

Table~\ref{tab:versions} lists the arXiv version of each corpus paper from which we extracted results. The first 17 papers were accessed on 24 September 2026 and rechecked on 25 September; the remaining 11 were accessed on 25 September.

\begin{table}[h]
\centering
\small
\begin{tabular}{@{}l l c l@{}}
\toprule
\textbf{Paper} & \textbf{arXiv ID} & \textbf{Version} & \textbf{Version date} \\
\midrule
\citet{li2026edge} & 2609.22753 & v1 & 19 Sep 2026 \\
\citet{li2026sixg} & 2609.23136 & v1 & 19 Sep 2026 \\
\citet{cheng2026thisthat} & 2609.23886 & v1 & 20 Sep 2026 \\
\citet{ren2026callscreen} & 2609.23959 & v1 & 21 Sep 2026 \\
\citet{jiang2026jevmem} & 2609.23986 & v1 & 21 Sep 2026 \\
\citet{rafe2026crash} & 2609.24052 & v1 & 21 Sep 2026 \\
\citet{robitza2026jevqa} & 2609.24395 & v1 & 21 Sep 2026 \\
\citet{ibrahim2026css} & 2609.24574 & v2 & 23 Sep 2026 \\
\citet{deng2026science} & 2609.24965 & v2 & 23 Sep 2026 \\
\citet{dagli2026fractal} & 2609.25498 & v2 & 24 Sep 2026 \\
\citet{yu2026visualjev} & 2609.25845 & v1 & 22 Sep 2026 \\
\citet{wu2026reflex} & 2609.26532 & v1 & 22 Sep 2026 \\
\citet{li2026judge} & 2609.26550 & v1 & 22 Sep 2026 \\
\citet{sun2026optionname} & 2609.26758 & v2 & 23 Sep 2026 \\
\citet{ma2026jevstar} & 2609.27331 & v1 & 23 Sep 2026 \\
\citet{li2026kite} & 2609.27535 & v1 & 23 Sep 2026 \\
\citet{huang2026radiology} & 2609.27607 & v1 & 23 Sep 2026 \\
\citet{zhang2026contract} & 2609.27678 & v1 & 23 Sep 2026 \\
\citet{ye2026numericjev} & 2609.28587 & v1 & 23 Sep 2026 \\
\citet{wu2026hijacking} & 2609.28613 & v1 & 23 Sep 2026 \\
\citet{abbasi2026harness} & 2609.28919 & v1 & 24 Sep 2026 \\
\citet{barbosa2026pentest} & 2609.28940 & v1 & 24 Sep 2026 \\
\citet{zhou2026pixeljev} & 2609.29283 & v1 & 24 Sep 2026 \\
\citet{guo2026justaskjev} & 2609.29429 & v1 & 24 Sep 2026 \\
\citet{rao2026rubricjudge} & 2609.29769 & v1 & 24 Sep 2026 \\
\citet{zhang2026jevmobile} & 2609.30186 & v1 & 24 Sep 2026 \\
\citet{ling2026wild} & 2609.30216 & v1 & 24 Sep 2026 \\
\citet{xu2026jevout} & 2609.30243 & v1 & 24 Sep 2026 \\
\bottomrule
\end{tabular}
\caption{arXiv versions of the 28 corpus papers used in this review.}
\label{tab:versions}
\end{table}

\section{Checklist Coding}
\label{app:coding}

We scored nine checklist items for the 27 corpus papers that evaluate a model; the ecosystem study~\citep{ling2026wild} is not scored. Each item was coded in two steps, applicability and then compliance. Cases that could not be determined from the paper are marked unclear and excluded from both counts.

\begin{description}
\item[B1] Applies to all 27. Compliant if any baseline reads label log-probabilities from an ordinary or same-backbone LM.
\item[B2] Applies only to papers that pass decisions to another model according to a confidence threshold, including routing that keeps low-confidence decisions on a stronger model; hand-offs on a fixed sample, on a schedule, or through an explicit option such as done/blocked do not count. Compliant if a cheap generative cascade is evaluated as a baseline.
\item[B3] Applies to papers that evaluate a fixed decision task on a labeled dataset large enough to train a task-specific model. Compliant if such a model (encoder, classical classifier, reward model or linear probe) is compared.
\item[T2] Applies to all 27. Compliant if the correctness or agreement of individual decisions across repeated, reordered or reworded requests is reported.
\item[T3] Applies to all 27. Compliant if option names are swapped across rubrics, or neutral and semantic names are compared.
\item[R2] Applies to all 27. Compliant if headline results carry confidence intervals or significance tests.
\item[C1] Applies to papers that use the TDM's probabilities or confidence in decisions, or describe them as calibrated. Compliant if calibration is measured (ECE, Brier score, reliability analysis or equivalent).
\item[C3] Applies to papers that set confidence thresholds for acceptance or escalation. Compliant if the thresholds were fitted on data separate from the evaluation set, or fixed before evaluation and not tuned on it.
\item[C4] Applies to papers that report TDM latency or claim a latency advantage for it. Compliant if latency is reported both with and without caching.
\end{description}

Items T1, T4, R1, R3 and C2 are disclosures that we could not score reliably from the text and are left unscored. For Finding~6, a paper counts as making choices after seeing evaluation data if it states, or its version history shows, that a design, recipe, metric or threshold was chosen after evaluation results were known; analyses disclosed as exploratory are not counted. Separately, a paper counts as adding comparators after seeing results if it states that a baseline or comparator was added once results were known; describing a comparison only as a follow-up is not enough. An internal numerical inconsistency is the same quantity reported with conflicting values within one version, excluding rounding.

\begin{table}[H]
\centering
\scriptsize
\setlength{\tabcolsep}{4pt}
\begin{tabular}{@{}l ccc ccc ccc@{}}
\toprule
\textbf{Paper} & \textbf{B1} & \textbf{B2} & \textbf{B3} & \textbf{T2} & \textbf{T3} & \textbf{R2} & \textbf{C1} & \textbf{C3} & \textbf{C4} \\
\midrule
\citet{cheng2026thisthat} & ? & -- & \xmark & \xmark & \xmark & \xmark & \cmark & -- & \xmark \\
\citet{ren2026callscreen} & \cmark & -- & \cmark & \cmark & \xmark & \cmark & \cmark & \cmark & \cmark \\
\citet{yu2026visualjev} & \cmark & -- & \xmark & \cmark & \xmark & \xmark & \cmark & \cmark & \cmark \\
\citet{zhou2026pixeljev} & \cmark & -- & \cmark & \cmark & \xmark & \cmark & \cmark & -- & \xmark \\
\citet{dagli2026fractal} & \xmark & -- & \cmark & \cmark & \xmark & \xmark & \xmark & -- & \xmark \\
\citet{li2026edge} & \xmark & -- & ? & \cmark & \xmark & \xmark & -- & -- & \cmark \\
\citet{li2026sixg} & \xmark & -- & -- & \cmark & \xmark & \xmark & -- & -- & \xmark \\
\citet{deng2026science} & \xmark & -- & -- & \cmark & \xmark & \xmark & -- & -- & \xmark \\
\citet{robitza2026jevqa} & \xmark & -- & \cmark & \xmark & \xmark & \xmark & \cmark & -- & \xmark \\
\citet{zhang2026contract} & \xmark & -- & \xmark & \cmark & \xmark & \cmark & -- & -- & \xmark \\
\citet{ye2026numericjev} & \xmark & -- & ? & \cmark & \xmark & \cmark & \xmark & -- & -- \\
\citet{zhang2026jevmobile} & \xmark & -- & -- & \xmark & \xmark & \xmark & -- & -- & \xmark \\
\citet{jiang2026jevmem} & \xmark & -- & -- & \xmark & \xmark & \xmark & \xmark & ? & \xmark \\
\citet{ma2026jevstar} & \xmark & -- & -- & \xmark & \xmark & \xmark & -- & -- & \xmark \\
\citet{wu2026reflex} & \xmark & \cmark & -- & ? & \xmark & \cmark & \cmark & ? & -- \\
\citet{abbasi2026harness} & \xmark & \xmark & -- & \xmark & \xmark & \xmark & \xmark & \cmark & \xmark \\
\citet{li2026judge} & \xmark & \xmark & \cmark & \cmark & \xmark & \cmark & \cmark & \cmark & \xmark \\
\citet{rao2026rubricjudge} & \xmark & \xmark & \xmark & \cmark & \xmark & \cmark & \cmark & \cmark & \xmark \\
\citet{huang2026radiology} & \xmark & -- & \xmark & \cmark & \xmark & \cmark & \cmark & -- & \xmark \\
\citet{guo2026justaskjev} & \xmark & -- & \cmark & \cmark & \xmark & \cmark & \cmark & -- & \xmark \\
\citet{barbosa2026pentest} & \xmark & -- & -- & \xmark & \xmark & \xmark & ? & ? & \xmark \\
\citet{ibrahim2026css} & \cmark & \xmark & \cmark & \xmark & \xmark & \cmark & \cmark & \cmark & -- \\
\citet{rafe2026crash} & \xmark & -- & \xmark & \cmark & \xmark & \cmark & \cmark & \cmark & \xmark \\
\citet{li2026kite} & \xmark & -- & ? & \xmark & \xmark & \cmark & \cmark & -- & ? \\
\citet{sun2026optionname} & \xmark & -- & -- & \cmark & \cmark & \cmark & -- & -- & -- \\
\citet{xu2026jevout} & \cmark & -- & -- & \cmark & \xmark & \cmark & -- & -- & -- \\
\citet{wu2026hijacking} & \xmark & -- & -- & \cmark & \xmark & \cmark & -- & -- & -- \\
\midrule
Compliant / applicable & 5/26 & 1/5 & 7/13 & 17/26 & 1/27 & 15/27 & 13/17 & 7/7 & 3/20 \\
\bottomrule
\end{tabular}
\caption{Per-paper checklist coding. \cmark: compliant; \xmark: applicable but not compliant; ?: unclear; --: not applicable.}
\label{tab:coding}
\end{table}

\end{document}